\documentclass[journal]{IEEEtran}
\makeatletter
\newcommand*{\rom}[1]{\expandafter\@slowromancap\romannumeral #1@}
\makeatother
\IEEEoverridecommandlockouts
\ifCLASSINFOpdf
\else
\fi

\usepackage[cmex10]{amsmath}
\usepackage{graphicx}
\usepackage{float,wrapfig}
\usepackage{multicol}
\usepackage[T1]{fontenc}
\usepackage{bm, tikz}
\usepackage[most]{tcolorbox}
\usepackage{tabularx,adjustbox,longtable}
\usepackage{subcaption}  
\usepackage{stfloats}
\usepackage{float}
\usepackage{subfloat}
\usepackage{caption}
\usepackage{graphics}
\usepackage{multicol}
\usepackage{lipsum}
\usepackage{microtype}
\usepackage{xcolor}
\usepackage{mathrsfs}
\usepackage{setspace}
\usepackage{lettrine}
\usepackage{newclude}
\usepackage[normalem]{ulem}
\usepackage{latexsym}
\usepackage{algpseudocode}
\usepackage{algorithm,algpseudocode}
\usepackage{algorithmicx}
\usepackage{multirow}
\usepackage{rotating}
\usepackage{booktabs}
\usepackage{wasysym}
\usepackage{mathrsfs}
\usepackage{bbm} 
\usepackage{filecontents}
\usepackage{adjustbox}

\usepackage{optidef}

\newtheorem{theorem}{Theorem}

\newtheorem{definition}{Definition}
\newtheorem{property}{Property}

\DeclareMathOperator{\R}{\mathbb{R}}

\DeclareMathOperator{\N}{\mathbb{N}}

\DeclareMathOperator{\sube}{\subseteq}

\DeclareMathOperator{\vep}{\varepsilon}

\DeclareMathOperator{\mD}{\mathcal{D}}

\DeclareMathOperator{\bW}{\mathbf{W}}

\DeclareMathOperator{\bx}{\mathbf{x}}
\DeclareMathOperator{\by}{\mathbf{y}}

\DeclareMathOperator{\bu}{\mathbf{u}}

\DeclareMathOperator{\bv}{\mathbf{v}}
\DeclareMathOperator{\ba}{\mathbf{a}}

\DeclareMathOperator{\bmu}{\bm{\mu}}

\ifCLASSOPTIONcompsoc
\else

\fi
\def\BibTeX{{\rm B\kern-.05em{\sc i\kern-.025em b}\kern-.08em
    T\kern-.1667em\lower.7ex\hbox{E}\kern-.125emX}}
\usepackage{amsmath,epsfig,cite,amsfonts,amssymb,psfrag,color}
\usepackage{epstopdf}

\def\BibTeX{{\rm B\kern-.05em{\sc i\kern-.025em b}\kern-.08em
    T\kern-.1667em\lower.7ex\hbox{E}\kern-.125emX}}

\def\hbar{\bar{\boldsymbol{h}}}

\begin{document}
\raggedbottom



\title{EMForecaster: A Deep Learning Framework for Time Series Forecasting in Wireless Networks with Distribution-Free Uncertainty Quantification}

\author{Xavier Mootoo, ~Hina~Tabassum, {\em Senior Member IEEE}, and Luca Chiaraviglio, {\em Senior Member IEEE} \vspace{-5mm}
 \thanks{X. Mootoo and H.~Tabassum are with the Department of Electrical Engineering and Computer Science, York University, Toronto, ON, Canada (E-mail: xmootoo@my.yorku.ca,  hinat@yorku.ca) L. Chiaraviglio is with the Department of Electronic Engineering, University of Rome Tor Vergata, Rome, Italy (E-mail: luca.chiaraviglio@uniroma2.it).}
}

\maketitle

\begin{abstract}
With the recent advancements in wireless technologies, forecasting electromagnetic field (EMF) exposure has become increasingly critical to enable proactive network spectrum and power allocation, as well as network deployment planning. In this paper, we develop a deep learning (DL)-empowered time series forecasting framework referred to as \textit{EMForecaster}. The proposed DL architecture employs patching to process temporal patterns at multiple scales, complemented by reversible instance normalization and mixing operations along both temporal and patch dimensions for efficient feature extraction. We then augment {EMForecaster} with a conformal prediction mechanism, which is independent of the data distribution, to enhance the trustworthiness of model predictions through uncertainty quantification of forecasts. In particular, the conformal prediction mechanism ensures that the ground truth lies within a prediction interval with target error rate $\alpha$, where $1-\alpha$ is referred to as coverage. However, a trade-off exists, as increasing coverage often results in wider prediction intervals. To address this challenge, we propose a new metric referred to as \textit{Trade-off Score}, that balances the trustworthiness of the forecast (i.e., coverage) and the width of prediction interval. Our empirical evaluation demonstrates that EMForecaster achieves superior performance across diverse EMF datasets, spanning both short-term and long-term prediction horizons. In point forecasting tasks, EMForecaster substantially outperforms current state-of-the-art DL approaches, showing improvements of 53.97\% over the Transformer architecture and 38.44\% over the average of all baseline models. 
In terms of conformal prediction performance, EMForecaster exhibits excellent balance between prediction interval width and coverage, as measured by the coverage-width tradeoff score. This balance is comparable to DLinear's performance while showing marked improvements of 24.73\% over the average baseline and 49.17\% over the Transformer architecture.  

\end{abstract}

\raggedbottom
\section{Introduction}

Amid rapid advancements in wireless technology, concerns are growing about the potential increase in electromagnetic field (EMF) exposure due to newer radio transmission frequencies, massive radiating elements, and the dense deployment of network infrastructure in fifth-generation (5G) and beyond (B5G) \cite{amin, 9518367}. EMF exposure can induce thermal effects, such as heating of the exposed tissues and organs, depending on the level of radiation absorbed. Consequently, regulatory authorities such as International Commission on Non-Ionizing Radiation Protection (ICNIRP) define rigorous limits on the maximum radio-frequency  EMF exposure between 100~kHz and 300~GHz \cite{ICNIRP}, which are adopted by many nations across the world. ICNIRP guidelines initially published in 1998 have been revised in 2020 to reflect recent advancements in scientific understanding of EMF exposure effects. In addition to the ICNIRP guidelines, several countries adopt more stringent EMF regulations \cite{stam2018comparison}. 


EMF monitoring and forecasting is thus becoming critical alongside the wireless technological advancements to ensure regulatory compliance, \textit{perform proactive network deployment, spectrum allocation, and power management}, and  address public health concerns. By providing insights into long-term EMF trends, reliable EMF forecasting enables proactive risk management and informed decision-making, thereby balancing wireless network innovation with safety and public trust. EMF forecasting capabilities allow network operators to optimize infrastructure placement and power settings while maintaining EMF levels within regulatory limits, and enable public health agencies to better assess population exposure patterns and implement preventive measures where necessary.

\subsection{Background Work}

To date, a variety of research studies have focused on  EMF exposure modeling and resource optimization in cellular networks using  analytical or simulation-based methods. For instance, stochastic geometry tools such as Poisson Point Processes (PPP) and shot-noise processes have been considered to model the spatial distribution of EMF sources and analyze their impact on EMF exposure  and coverage in cellular networks \cite{app10238753, gontier2024impactdynamicbeamformingemf, 9511258,  gontier2024uplinkdownlinkemfexposure,  10536047, 10504892, 10047969, 9462948, 10225716}. On the other hand, several studies have focused on minimizing EMF exposure in cellular systems  \cite{sambo2016electromagnetic},\cite{jiang2023rate}.  Resource management schemes, including user-scheduling \cite{sambo2014user}, spectrum allocation \cite{sambo2016electromagnetic}, multi-cell scheduling \cite{sambo2017electromagnetic}, user association \cite{matalatala2018joint}, beamforming optimization \cite{wang2011evaluation, ying2013beamformer}, and cross-layer protocols \cite{penhoat2015enabling, diez2015reducing} have also been considered to minimize EMF exposure. Network-based approaches considered minimizing EMF exposure by optimizing  cellular network planning, as discussed in \cite{chiaraviglio2018planning, ITU:2019, matalatala2019multi, matalatala2018optimal, amaldi2003planning, oughton2019open}.

Nevertheless, the aforementioned model-based approaches often suffer from inaccuracies due to necessary modeling assumptions, making them less applicable in complex real-world environments. Developing analytical models for such scenarios is not only intractable, but also computationally demanding, as they result in intricate mathematical expressions. Moreover, these methods are constrained by specific wireless channel assumptions, limiting their applicability to dynamic and unpredictable network conditions. Additionally, they lack the ability to learn from historical patterns or trends, preventing them from effectively performing proactive network resource management and planning. As a result, their practicality in real deployments remains a significant challenge.

Recently, deep learning (DL) approaches have gained significant attention in time series forecasting, an area historically dominated by statistical models and traditional machine learning methods \cite{miller2024survey}.  Traditional techniques, such as Autoregressive Integrated Moving Average (ARIMA) \cite{arima} combines differencing, autoregression, and moving average components to model linear relationships in stationary time series data \cite{arima}, and has long been a standard benchmark in time series modeling, often outperforming more modern methods in standardized testing environments. 
However, the advent of sophisticated DL architectures, such as transformers, has begun to shift the paradigm \cite{miller2024survey}. DL methods offer distinct advantages over conventional forecasting methods. These models can learn complex nonlinear relationships without explicit feature engineering and adapt to dynamic changes by modeling hierarchical structures. Unlike traditional methods that require manual pre-processing such as seasonal decomposition, DL models can process raw time series data directly while handling multiple input variables and their interactions. 

 DL methods are increasingly being adopted for EMF pattern analysis and prediction due to their ability to handle complex, nonlinear relationships in the data.  In \cite{kiouvrekis2024comparative}, Kiouvrekis \textit{et al.} employed hierarchical clustering to analyze EMF measurements across 205 schools in Thessaly, Greece. The findings reveal that EMF exposure patterns were independent of urban population density in the 27 MHz–3 GHz range. Bakcan \textit{et al.} integrated static electric field measurements with artificial neural networks (ANN) in \cite{bakcan2022measurement} to predict EMF exposure at unmeasured locations across a campus setting, validating results against Information and Communication Technology Agency (ICTA) and ICNIRP standards. Focusing on temporal prediction, Pala \textit{et al.} in \cite{pala2021examining} analyzed EMF dataset comprised of 60 monthly mean values in the range of 1~Hz - 400~kHz using the Wavecontrol SMP2 device, and demonstrated that long short-term memory (LSTM) outperforms recurrent neural network (RNN) and traditional statistical approaches. 
However, the model was based on a very limited dataset for both training and testing, raising concerns about overfitting.
{Moreover, the study did not consider fundamental time series parameters such as \textit{lookback window} (how much historical data to use), \textit{forecast horizon} (how far ahead to predict), and \textit{rolling window stride} (how to slide the prediction window through time).} Furthermore, the dataset’s frequency range was significantly low to encompass modern wireless networks. 

Very recently, Nguyen \textit{et al.} in \cite{nguyen2024deep} demonstrated the effectiveness of Transformers \cite{transformer} over the core DL  architectures, including the multilayer perceptron (MLP), LSTM, and convolutional neural networks (CNNs) for short-term EMF forecasting using the data provided in \cite{kurnaz2020rfemf}.  The training data was collected from real measurements taken in a city of Ordu, Turkey \cite{kurnaz2020rfemf}, considering multi-step input and output sequences. 
{Building upon the pioneering application of DL to EMF time series forecasting in \cite{nguyen2024deep}, we note that the framework is restricted to short-term prediction windows (maximum duration of 20 minutes). In addition, although the dataset was carefully collected with 24-hour recordings from 17 sites at 15-second intervals, it is relatively modest by DL standards, where larger datasets typically enable more robust performance evaluation. The limited temporal scope of the data presents challenges in capturing multi-day patterns and long-term exposure trends, as the resulting non-stationary time series  limits model predictive capabilities. Furthermore, our analysis in Section~V reveals high correlation in the data, suggesting limited diversity in EMF exposure scenarios, which may impact the ability to assess each model's generalizability. Moreover, the work employed classical DL architectures, not specifically designed for temporal modeling. Finally, although hyperparameter values were documented, the absence of ablation studies and sensitivity analyses makes it difficult to fully understand each model's behavior and validate design choices.}


\subsection{Contributions}
None of the existing research has developed a reliable DL-empowered time series forecasting framework for wireless cellular network applications in general, nor specifically for EMF forecasting. 
The black-box nature of DL models makes it difficult to interpret predictions or understand the underlying decision-making process, raising concerns in applications where transparency and reliability are paramount.  
To this end, our contributions are listed as follows:

$\bullet$ {We develop a novel DL-driven time series forecasting solution, referred to as \textit{EMForecaster}, to forecast EMF patterns considering both long-term and short-term time series, across a variety of locations with a forecast horizon of up to 50 hours. EMForecaster exploits several modern DL techniques such as Reversible Instance Normalization (RevIN), which minimizes the impact of distribution shifts or non-stationarity in the EMF data, in addition to a patching and patch-embedding module. EMForecaster then applies the spatiotemporal backbone (STB) to the patch-embedded data where we employ a mixing operation on the learned patch representations, enabling learning of hierarchical patterns at multiple scales.}


$\bullet$  We augment the \textit{EMForecaster} with a CP framework to ensure that the ground truth
lies within a certain prediction interval with \textit{error rate} $\alpha$. The performance is measured in terms of the independent coverage (IC) of each individual forecast point, the joint coverage (JC) of the forecast points across the  forecasted window, and mean prediction interval width (MIW). Unlike traditional uncertainty quantification methods, which often depend on strong distributional assumptions, CP is agnostic to the original data distribution, making it compatible with any DL model and dataset.  
  
$\bullet$ We propose the Trade-off Score (TOS)\textemdash a unified metric that evaluates the effectiveness of CP methods by considering two key aspects: (1) how often the true values fall within the predicted ranges (coverage) and (2) how wide these predicted ranges are (interval width). TOS quantifies how well a method balances this fundamental trade-off. While wider prediction ranges are more likely to contain the true values, they provide less precise and thus less useful forecasts. A higher TOS indicates better performance, as the true values fall within the prediction range without requiring excessively wide intervals.

$\bullet$ {Our experiments, covering a wide variety of environments and forecast horizons, display EMForecaster's capabilities over the existing DL models.} In point forecasting, EMForecaster demonstrates improvements of 53.97\% over the Transformer architecture and 38.44\% over the average of all baseline models. When compared to the DLinear architecture, EMForecaster maintains  8.01\% gains. For conformal forecasting, EMForecaster provides an excellent balance between prediction interval coverage and width, as measured by the TOS. This balance is comparable to DLinear's performance while showing marked improvements of 24.73\% over the average baseline and 49.17\% over the Transformer architecture.  


$\bullet$ {We conduct a comprehensive EMF data analysis by applying spectral decomposition, stationarity testing, and spatial correlation analysis on both short-term and long-term EMF exposure datasets. Stationarity is particularly crucial, which implies that time windows are reasonably independently and identically distributed (i.i.d.), therefore satisfying the exchangeability condition necessary for conformal prediction (see Section~II.C for more details).
}


\subsection{Paper Organization}
{The remainder of this paper is organized as follows. Section~\ref{sec:background} introduces fundamental concepts in time series forecasting. Section~\ref{sec:architecture} presents our proposed DL model, EMForecaster, followed by conformal prediction for uncertainty quantification in Section~\ref{sec:conformal}. Section~\ref{sec:experiments} details the datasets, experimental setup, and comprehensive data analysis. Section~\ref{sec:results} presents the baseline models, results, and discussion. Finally, Section~\ref{sec:conclusion} concludes the paper.}


\section{Fundamentals of Time Series Forecasting \label{sec:background}}
In this section, we discuss the fundamental concepts related to time series forecasting, such as stationarity and periodicity as well as the concepts related to building trustworthy forecasting such as exchangeability and conformal prediction.

\subsection{General Definitions}
Let $(x_t)_{t = 1}^T$ be a time series of total length $T$. Given a subsequence (or window) $\bx = (x_t)_{t=K}^{K+L}$ with starting index $K$ and sequence length $L$, the goal of time series forecasting is to predict its continuation ${\by} = (x_{t})_{t=L+K+1}^{K+L+O}$, where $O$ is the length of the forecast horizon. We will typically view $\bx \in \R^{L}$ and ${\by} \in \R^{O}$ as vectors. When preprocessing the initial time series $(x_t)_{t = 1}^T$, we use the sliding window technique, which samples windows $(x_t)_{t=1}^{L}, (x_t)_{t=2}^{L+1}, \dots, (x_t)_{t = L-T}^{T}$ in which we ``slide" the starting index $K$ by $1$, and sample windows of length $L$ at each index. In general, this forms our dataset, while keeping training, validation, and test sets  temporally separated, i.e., they consist of mutually exclusive regions of the sub-sequence. Large values of $O$ represents long-term forecasting horizon and vice versa.

\subsection{Stationarity of Time Series}
Stationarity is an invariant property where statistical properties of a time series remain consistent over time.  Stationarity can be classified as: \textit{weak stationarity}, which only considers the covariance of a time series, and \textit{strict stationarity}, which assumes distributions remain invariant over time. While classical models such as ARIMA \cite{arima} explicitly require stationarity to maintain their statistical properties,  DL has introduced new perspectives on handling non-stationary data \cite{liu2022non}. Traditional approaches necessitate explicit transformation of non-stationary data through techniques such as differencing or seasonal decomposition, whereas DL models may learn and adapt to certain types of non-stationarity through hierarchical feature learning capabilities. However, empirical evidence suggests that even for DL models, highly non-stationary distributions can lead to suboptimal performance when compared to traditional models. Thus, characterizing the stationarity of a time series can improve the forecasting quality \cite{santoro2024comparison}. One standard method to test stationarity is the \textit{Augmented Dickey-Fuller (ADF) test}, which examines the presence of unit roots in time series using an autoregressive model estimated through ordinary least squares regression \cite{adf}:
\begin{equation}
{x_t = c + w_1 t + w_2 x_{t-1} + \sum_{i=1}^p \phi_i \Delta x_{t-i} + \varepsilon_t}
\end{equation}
where $x_t$ is the time series value at time $t$, $c$ is the estimated drift constant, $w_1$ is the estimated time trend coefficient capturing deterministic trends, $w_2$ is the estimated process root determining persistence, $\phi_i$ are the estimated autoregressive coefficients capturing short-term dynamics, $\Delta x_{t-i}$ represents lagged differences $(x_{t-i} - x_{t-i-1})$, and $\varepsilon_t$ is white noise with zero mean and constant variance. The test evaluates the hypothesis $H_0: w_2 = 1$ (non-stationary) against $H_a: w_2 < 1$ (stationary) with test statistic \cite{adf}:
\begin{equation}
\text{ADF} = \frac{\hat{w}_2 - 1}{\text{SE}(\hat{w}_2)}
\end{equation}
where $\hat{w}_2$ is the estimated value of $w_2$ and $\text{SE}(\hat{w}_2)$ is its standard error measuring estimation uncertainty. A more negative test statistic provides stronger evidence for stationarity, with the additional terms in the regression controlling for trends and autocorrelation to ensure robust testing. In our analysis provided in Figure~\ref{fig:adf}, ADF testing reveals distinct stationarity characteristics among different EMF datasets.


\subsection{Conformal Prediction and Exchangeability in Time Series 
}
The goal of CP is to provide a \emph{prediction region}, $\Gamma^{\alpha}$, for a given \textit{significance level (or error rate)} $\alpha \in (0,1)$, ensuring that the ground truth falls within this region with a probability of at least $(1 - \alpha)$. This framework's key advantage lies in its ability to produce valid uncertainty estimates for any black-box predictor, making it particularly valuable in domains where reliable uncertainty quantification is crucial \cite{enbpi, ctsf}. CP enables constructing prediction intervals with guaranteed finite-sample coverage under minimal assumptions on the data distribution, requiring only exchangeability of the data.

Exchangeability is a key statistical property where the order of time observations does not affect their joint probability\textemdash similar to how shuffling a well-mixed deck of cards does not change the probability of drawing any particular sequence. This property underpins CP's theoretical guarantees, but poses challenges for time series analysis. Time series data inherently violates exchangeability since future observations can depend on past states. In wireless networks, for instance, network utilization at time $t$ can depend on its state at $t-1$. To address this, several adaptations have been proposed, notably in \cite{ctsf}, where the strict exchangeability assumption is relaxed through locally exchangeable windows. This approach preserves CP's validity while acknowledging the temporal nature of the data by treating similar historical patterns as locally exchangeable within defined contexts.

\subsection{Fast-Fourier Transform and Periodicity}
The Fast Fourier Transform (FFT) decomposes complex temporal patterns into their constituent frequency components, enabling efficient characterization of cyclical behaviors within the data \cite{bloomfield2004fourier}. Mathematically, the FFT efficiently computes the Discrete Fourier Transform (DFT), which decomposes a time series $\bx = (x_t)_{t = 1}^T$ into its frequency components by:
\begin{equation}
X_k = \sum_{t=1}^T x_t e^{-2\pi i k t/T}, \quad k = 0,\ldots,{T}-1
\end{equation}
where the amplitude $|X_k|$ reveals the  dominant frequency components (cyclical patterns) in the time series. By transforming time-domain signals into the frequency domain, FFT analysis reveals both the presence and strength of periodic patterns. This spectral decomposition proves particularly valuable in scenarios where multiple overlapping cycles may exist.

\section{State-of-the-Art DL Models for Time Series Forecasting: A Review}
In this section, we provide a review of existing DL models for time-series forecasting and summarized their benefits and drawbacks in \textbf{Table~1}. The described DL models have been applied for a variety of wireless applications to date including EMF prediction \cite{kiouvrekis2024comparative, bakcan2022measurement, pala2021examining, nguyen2024deep}, network traffic prediction \cite{nikravesh2016mobile, di2023multivariate, dalgkitsis2018traffic, habib2024transformer, perez2024dissecting, isravel2024multivariate, sone2020wireless}, channel prediction \cite{zhang2021deep, sone2020wireless},  and network quality-of-service (QoS) prediction \cite{hameed2022toward, colpitts2023short}.

\begin{table*}[ht]
\centering
\caption{Comparison of Deep Learning Models for Time Series Forecasting in Wireless Networks}
\small
\begin{tabular}{p{1.7cm}p{7.5cm}p{7.5cm}}
\toprule
\textbf{Model} & \textbf{Pros} & \textbf{Cons} \\
\midrule
MLP \cite{kiouvrekis2024comparative, bakcan2022measurement, nikravesh2016mobile, di2023multivariate, nguyen2024deep} & 
Simple architecture; Fewer parameters; Fast training/inference; Low computational cost; Adaptable to different inputs & 
Limited temporal modeling; Poor with long-range dependencies; Requires feature engineering; Overfits on complex EMF patterns; Cannot capture periodic patterns \\
\midrule
CNN \cite{sone2020wireless, ak2021forecasting, zhou2022deep, di2023multivariate, nguyen2024deep} & 
Captures local patterns effectively; Efficient parameter sharing; Detects signal hierarchies; Noise-robust; Parallel processing & 
Limited receptive field; Issues with irregular sampling; Weak global context modeling; Fixed kernel limitations; Requires careful design \\
\midrule
LSTM \cite{pala2021examining,  dalgkitsis2018traffic, zhang2021deep, colpitts2023short, di2023multivariate, isravel2024multivariate, nguyen2024deep} & 
Designed for long-term dependencies; Memory cell preserves patterns; Effective gating mechanisms; Handles variable sequences; Models complex dynamics & 
Sequential processing limits parallelization; Computationally expensive for long sequences; Gradient issues; Many parameters; Difficult interpretation \\
\midrule
Transformer \cite{transformer, hameed2022toward, habib2024transformer, nguyen2024deep} & 
Models dependencies at any distance; Parallel computation; Global context modeling; Multi-head attention; Scales to long sequences & 
Quadratic complexity; High memory usage; Requires positional encoding; No inherent sequential understanding; Weak with fine-grained patterns \\
\midrule
PatchTST \cite{patchtst, perez2024dissecting} & 
Captures semantic information through patches; Channel independence reduces overfitting; Efficient multi-channel processing; Better noise handling; Superior time series performance & 
Sensitive to patch parameters; May lose fine-grained details; Parameter tuning challenges; Higher complexity than linear models; More complex implementation \\
\midrule
DLinear \cite{dlinear, perez2024dissecting} & 
Effective seasonal-trend decomposition; Computationally efficient; Outperforms many Transformer models; Resistant to overfitting; Simpler architecture & 
Limited non-linear modeling; Fixed decomposition constraints; Lower expressiveness; Critical hyperparameter selection; Struggles with non-stationary data  \\
\midrule
\textbf{EMForecaster} (Proposed) & 
Patching with spatiotemporal mixing operations; Enhanced feature extraction with reversible normalization; Superior performance across diverse datasets; Balanced prediction intervals for conformal prediction; Effective for both short/long-term forecasting; Low computational cost & 
More complex architectural design; Requires deeper understanding of the model; More challenging initial implementation; Requires adequate volume of data to generalize \\
\bottomrule
\end{tabular}
\end{table*}

\subsection{Multilayer Perceptron (MLP)}
A Multilayer Perceptron (MLP) is a type of feedforward neural network consisting of an input layer, one or more hidden layers, and an output layer, where each layer is fully connected to the next. The MLP can model complex, non-linear relationships by applying a non-linear activation function \(\sigma\) at each node in the hidden and output layers. Given an input time series \(\mathbf{x} \in \mathbb{R}^L\), the transformation at the \(k\)-th hidden layer is defined as 
\[
\mathbf{h}^{(k)} = \sigma\left( \mathbf{W}^{(k)} \mathbf{h}^{(k-1)} + \mathbf{b}^{(k)} \right),
\]
where each \(\mathbf{W}^{(k)}\) is a learnable weight matrix and \(\mathbf{b}^{(k)}\) is a learnable bias vector for layer \(k\), with \(\mathbf{h}^{(0)} = \mathbf{x}\). The output layer computes the predicted forecast \(\hat{\mathbf{y}} \in \mathbb{R}^O\) as:
\[
\hat{\mathbf{y}} = \sigma_{\text{out}}\left( \mathbf{W}^{(K+1)} \mathbf{h}^{(K)} + \mathbf{b}^{(K+1)} \right),
\]
where \(K\) is the number of hidden layers, and \(\sigma_{\text{out}}\) is an activation function appropriate for the task, for example, softmax for time series classification or the identity function for time series forecasting. By stacking multiple layers and introducing nonlinearity, MLPs are capable of approximating complex functions, making them widely applicable in various machine learning and time series forecasting contexts.

\subsection{Convolutional Neural Network (CNN)}
A 1D Convolutional Neural Network (CNN) is a type of neural network designed to process sequential data, such as time series or one-dimensional signals, by learning spatial or temporal hierarchies through convolutional operations \cite{cnn1, cnn2, bai2018empirical}. In a 1D CNN, the input signal \(\mathbf{x} \in \mathbb{R}^L\) is convolved with a set of learnable filters (or kernels), producing feature maps that capture local patterns. The output of the \(j^{th}\) filter in layer \(k\) is given by:
\[
\mathbf{h}^{(k)}_j[n] = \sigma\left( \sum_{m=1}^{M_k} \mathbf{w}^{(k)}_{j}[m] \mathbf{h}^{(k-1)}[n-m+1] + b^{(k)}_j \right),
\]
where \(\mathbf{w}^{(k)}_j\) is the filter of size \(M_k\), \(b^{(k)}_j\) is the bias term, \(\sigma\) is a non-linear activation function, and \(n\) indexes the position in the sequence. The feature maps from one layer are passed to the next, allowing the network to progressively learn higher-level abstractions. At the output layer, these learned features can be fed into a fully connected layer or another downstream task. The use of convolutional layers makes 1D CNNs particularly effective at capturing local dependencies in sequential data while reducing the number of parameters compared to fully connected architectures.

\subsection{Long-Short Term Memory (LSTM) Networks}
The \textit{Long Short-Term Memory} (LSTM) network is a specialized recurrent neural network (RNN) designed to model long-term dependencies in sequential data by maintaining a memory cell \(\mathbf{c}_t\) at each time step \cite{rnn, gru, lstm}. This structure mitigates vanishing and exploding gradient problems, making LSTMs effective for univariate time series forecasting and classification.
At each time step $t$, a hidden state $\mathbf{a}_t \in \mathbb{R}^d$ serves as a summary of the information processed by the model up to time $t$ where $d$ is the hidden state dimension. The goal of $\ba_t$ is to capture both short-term patterns and contextual signals needed for immediate predictions. The memory cell $\mathbf{c}_t \in \mathbb{R}^d$, on the other hand, provides a more stable, long-term storage, selectively preserving information that might be important for future time steps. Together, the hidden state and memory cell allow the model to balance between remembering essential past information and responding to new inputs. Each gate (update, forget, and output) produces a vector $\Gamma \in \mathbb{R}^d$ that modulates the flow of information, ensuring that the model learns to focus on relevant parts of the sequence while discarding irrelevant or redundant details. This formulation enables the LSTM to handle both short-term dependencies (through $\mathbf{a}_t$) and long-term dependencies (through $\mathbf{c}_t$) effectively, which is crucial for univariate time series forecasting.

\textbf{Candidate memory cell and gates.} The LSTM computes a \textit{candidate memory cell} \(\tilde{\mathbf{c}}_t\) as follows:
\[
\tilde{\mathbf{c}}_t = \tanh\left( \mathbf{W}_c [\mathbf{a}_{t-1}, x_t] + \mathbf{b}_c \right)
\tag{1}
\]
Here, \([\mathbf{a}_{t-1}, x_t] \in \mathbb{R}^{d+1}\) denotes the concatenation of the previous hidden state \(\mathbf{a}_{t-1}\) and the current scalar input \(x_t\). The weight matrix \(\mathbf{W}_c \in \mathbb{R}^{d \times (d+1)}\) maps the concatenated vector to the hidden size \(d\), and \(\mathbf{b}_c \in \mathbb{R}^d\) is the bias vector.

The gates are computed as:
\begin{itemize}
    \item \textbf{Update Gate:} Controls how much of the candidate memory cell to incorporate into the current memory cell.
    \begin{align}
         \Gamma_u = \sigma\left( \mathbf{W}_u [\mathbf{a}_{t-1}, x_t] + \mathbf{b}_u \right)
    \end{align}
    where \(\mathbf{W}_u \in \mathbb{R}^{d \times (d+1)}\) and \(\mathbf{b}_u \in \mathbb{R}^d\).

    \item \textbf{Forget Gate:} Determines how much of the previous memory cell to retain.
    \begin{align}
         \Gamma_f = \sigma\left( \mathbf{W}_f [\mathbf{a}_{t-1}, x_t] + \mathbf{b}_f \right)
    \end{align}
    where \(\mathbf{W}_f \in \mathbb{R}^{d \times (d+1)}\) and \(\mathbf{b}_f \in \mathbb{R}^d\).

    \item \textbf{Output Gate:} Controls the prominence of the new activation.
    \begin{align}
        \Gamma_o = \sigma\left( \mathbf{W}_o [\mathbf{a}_{t-1}, x_t] + \mathbf{b}_o \right)
    \end{align}
    where \(\mathbf{W}_o \in \mathbb{R}^{d \times (d+1)}\) and \(\mathbf{b}_o \in \mathbb{R}^d\).
\end{itemize}

\textbf{Memory and activation updates}. The memory cell update and new hidden state (activation) are computed as:
\begin{align}
    \mathbf{c}_t &= \Gamma_u \odot \tilde{\mathbf{c}}_t + \Gamma_f \odot \mathbf{c}_{t-1} \\
    \mathbf{a}_t &= \Gamma_o \odot \tanh\left( \mathbf{c}_t \right)
\end{align}
Here, \(\odot\) denotes element-wise multiplication. Given a time series $\bx = (x_t)_{t = 1}^L$, we first obtain the entire sequence of hidden states \(\{\mathbf{a}_1, \mathbf{a}_2, \dots, \mathbf{a}_L\}\) recursively. A common approach is to use the final hidden state:
\begin{align}
    \mathbf{a}_{\text{final}} = \mathbf{a}_L \in \mathbb{R}^{d}
\end{align}
This representation is then passed through a fully connected layer to generate the forecast:
\begin{align}
    \hat{y} = \mathbf{w}_{\text{out}}^T \mathbf{a}_{\text{final}} + b_{\text{out}}
\end{align}
where \(\mathbf{w}_{\text{out}} \in \mathbb{R}^{d}\) and \(b_{\text{out}} \in \mathbb{R}\) produce a scalar output \(\hat{y} \in \mathbb{R}\) for the one-step-ahead forecast.
For multi-step forecasting in the univariate case, we can either: (1) use a single output neuron and apply the model recursively, feeding each prediction back as input; or (2) use multiple output neurons to directly predict the entire forecast horizon:
\begin{align}
    \hat{\mathbf{y}} = \mathbf{W}_{\text{out}} \mathbf{a}_{\text{final}} + \mathbf{b}_{\text{out}}
\end{align}
where \(\mathbf{W}_{\text{out}} \in \mathbb{R}^{O \times d}\) and \(\mathbf{b}_{\text{out}} \in \mathbb{R}^O\) produce an output \(\hat{\mathbf{y}} \in \mathbb{R}^O\), with \(O\) being the length of the forecast horizon.

\subsection{The Transformer}

\textbf{The Transformer architecture.} The \textit{Transformer} is a versatile model architecture that uses self-attention mechanisms to capture dependencies across all time steps in a sequence, regardless of their temporal distance \cite{transformer}. This makes it particularly suitable for time series forecasting, where both short-term and long-term patterns can be critical. Unlike recurrent models such as LSTMs, which process sequences step-by-step, the Transformer architecture processes the entire input sequence in parallel, enabling direct interactions between all positions through its \textit{self-attention} mechanism. In this section, we focus on the \textit{encoder-only} Transformer architecture for univariate time series forecasting.

\textbf{Input and positional encoding}. For a univariate time series with input sequence $\mathbf{x} \in \mathbb{R}^L$, where $L$ is the input sequence length, we first project each scalar value to the model's hidden dimension using a learned projection matrix $\mathbf{W}_\text{in} \in \mathbb{R}^{d \times 1}$. Since the Transformer lacks an inherent notion of temporal order, a \textit{positional encoding} vector $\mathbf{p}_t = (p_{t,1}, \dots, p_{t,d})^T \in \mathbb{R}^d$ is then added to each projected value, with each entry defined by:
\begin{equation}
    p_{t,i} = \begin{cases} 
    \cos(t/10000^{2i/d}) &\text{if $i$ is even}  \\ 
    \sin(t/10000^{2(i-1)/d}) &\text{if $i$ is odd}
    \end{cases}
\end{equation}
The full projection transformation for each time point $t$ is given by:
\begin{align}
\mathbf{z}_t = \mathbf{W}_\text{in} x_t + \mathbf{p}_t,
\end{align}
This produces an initial sequence representation $\mathbf{Z} = [\mathbf{z}_1, \mathbf{z}_2, \dots, \mathbf{z}_L]^T \in \mathbb{R}^{L \times d}$.

\textbf{The self-attention mechanism.} The core of the Transformer encoder is the self-attention mechanism, which allows each time step to attend to all other time steps in the sequence. For each embedded input $\mathbf{z}_t$, the model computes three vectors: the query $\mathbf{q}_t$, key $\mathbf{k}_t$, and value $\mathbf{v}_t$:
\begin{align}
\mathbf{q}_t &= \mathbf{W}_Q \mathbf{z}_t, \quad \mathbf{k}_t = \mathbf{W}_K \mathbf{z}_t, \quad \mathbf{v}_t = \mathbf{W}_V \mathbf{z}_t,
\end{align}
where $\mathbf{W}_Q, \mathbf{W}_K, \mathbf{W}_V \in \mathbb{R}^{d \times d}$ are learned projection matrices. The attention score between time steps $t$ and $s$ is computed as the scaled dot product:
\begin{align}
\alpha_{ts} = \frac{\mathbf{q}_t^T \mathbf{k}_s}{\sqrt{d}}.
\end{align}

These scores are normalized across all time steps using the softmax function:
\begin{align}
\alpha_{t1}, \alpha_{t2}, \dots, \alpha_{tL} = \text{softmax}\left( \frac{\mathbf{q}_t^T \mathbf{K}^T}{\sqrt{d}} \right),
\end{align}
where $\mathbf{K} = [\mathbf{k}_1, \mathbf{k}_2, \dots, \mathbf{k}_L]^T \in \mathbb{R}^{L \times d}$. The output of the attention mechanism for time step $t$ is the weighted sum:
\begin{align}
\text{Attention}(\mathbf{q}_t, \mathbf{K}, \mathbf{V}) = \sum_{s=1}^L \alpha_{ts} \mathbf{v}_s,
\end{align}
where $\mathbf{V} = [\mathbf{v}_1, \mathbf{v}_2, \dots, \mathbf{v}_L]^T \in \mathbb{R}^{L \times d}$.

\textbf{Encoder block design.} The Transformer encoder consists of $N$ identical blocks stacked sequentially. Each encoder block contains two main components: a multi-head self-attention layer and a position-wise feed-forward network, both wrapped with residual connections and layer normalization. The multi-head attention component enhances the model's capacity by allowing multiple attention mechanisms to operate in parallel:
\begin{align}
\text{MultiHead}(\mathbf{Z}) = \text{concat}(\mathbf{H}_1, \mathbf{H}_2, \dots, \mathbf{H}_h) \mathbf{W}_O,
\end{align}
where $h$ is the number of attention heads, $\mathbf{H}_i \in \mathbb{R}^{L \times d_h}$ is the output of the $i$-th head, $d_h = d/h$ is the dimensionality of each head, and $\mathbf{W}_O \in \mathbb{R}^{d \times d}$ is a learned projection matrix. Each attention head $\mathbf{H}_i$ is computed independently:
\begin{align}
\mathbf{H}_i = \text{Attention}(\mathbf{Z} \mathbf{W}_Q^{(i)}, \mathbf{Z} \mathbf{W}_K^{(i)}, \mathbf{Z} \mathbf{W}_V^{(i)}),
\end{align}
where $\mathbf{W}_Q^{(i)}, \mathbf{W}_K^{(i)}, \mathbf{W}_V^{(i)} \in \mathbb{R}^{d \times d_h}$ are the learned projection matrices specific to the $i$-th head.

Following multi-head attention, a position-wise \textit{feedforward network} (FFN) is applied independently at each time step:
\begin{align}
\text{FFN}(\mathbf{z}_t) = \mathbf{W}_2(\sigma(\mathbf{W}_1 \mathbf{z}_t + \mathbf{b}_1)) + \mathbf{b}_2,
\end{align}
where $\mathbf{W}_1 \in \mathbb{R}^{d_\text{ff} \times d}$, $\mathbf{W}_2 \in \mathbb{R}^{d \times d_\text{ff}}$, $\mathbf{b}_1 \in \mathbb{R}^{d_\text{ff}}$, $\mathbf{b}_2 \in \mathbb{R}^{d}$, and $d_\text{ff}$ is the hidden dimension of the FFN. The Transformer encoder block also includes \textit{layer normalization} and residual connections:
\begin{align}
\mathbf{Z}' &= \text{LayerNorm}(\mathbf{Z} + \text{MultiHead}(\mathbf{Z})), \\
\mathbf{Z}'' &= \text{LayerNorm}(\mathbf{Z}' + \text{FFN}(\mathbf{Z}')).
\end{align}

\textbf{Forecasting with Transformers.} For univariate time series forecasting with forecast horizon $O$, the final hidden representations $\mathbf{Z}_{\text{final}} = [\mathbf{z}''_1, \mathbf{z}''_2, \dots, \mathbf{z}''_L]^T \in \mathbb{R}^{L \times d}$ from the last encoder block capture both local and global temporal dependencies across the input sequence. To generate the forecast sequence, we use all time points by first flattening the final hidden representations and then applying a linear transformation:
\begin{align}
\mathbf{z}_{\text{flat}} &= \text{flatten}(\mathbf{Z}_{\text{final}}) \in \mathbb{R}^{Ld} \\
\hat{\mathbf{y}} &= \mathbf{z}_{\text{flat}} \mathbf{W}_{\text{out}} + \mathbf{b}_{\text{out}},
\end{align}
where $\mathbf{W}_{\text{out}} \in \mathbb{R}^{Ld \times O}$ and $\mathbf{b}_{\text{out}} \in \mathbb{R}^{O}$. This produces the forecast sequence $\hat{\mathbf{y}} = (\hat{y}_1,  \dots, \hat{y}_O]^T)^T \in \mathbb{R}^{O}$, which represents the predicted values for the next $O$ time steps following the input sequence.

\subsection{PatchTST}
\textbf{Patching}. Time series forecasting attempts to model the relationships between data points from different time steps. A single time point does not equivocate to the level of semantic meaning of words within a sentence, similar to how individual letters do not carry high semantic information by themselves. In contrast, subseries-level patches enhances the model and captures higher semantic information, similar to modeling the relationship of words or subwords in a sentence \cite{patchtst}. However, most Transformer-based methods for time series do not rely on patching and patch-wise attention, and instead treat individual time points as tokens \cite{logtrans, reformer, informer, autoformer, fedformer, pyraformer, triformer}. Instead of using individual time points as tokens, \textit{PatchTST} treats patches as tokens to be fed into the standard Transformer architecture which greatly improves forecasting performance \cite{transformer, patchtst}.

\subsection{DLinear}
DLinear is linear neural network that borrows ideas from statistical time series modelling, and applies seasonal-trend decomposition to the input time series $\bx$ before processing it through two distinct linear transformations \cite{dlinear}. While Transformer-based methods have been the most prevalent architecture for time series forecasting, they can often struggle with the temporal information loss due to their permutation-invariant self-attention mechanism. DLinear, however, has demonstrated superior performance over many state-of-the-art Transformer-based models across standardized benchmarks, questioning of the reliability of Transformer-based approaches and highlighting the effectiveness of more simpler deep learning approaches for time series forecasting.
The method for DLinear is as follows. Given a time series $\bx$, we first pad $\bx$ with $m$ values on both sides of the series using its first and last values, extending its length to $L+2m$. We then define the moving average $\bmu_t$ as:
\begin{align} \label{moving_avg}
\bmu_t = \frac{1}{2m+1} \sum_{k = t-m}^{t+m} \bx_{k}
\end{align}
for each $0 \leq t \leq L$, where $m \in \N$ is a hyperparameter. The local averaging window $m$ smooths the input time series to form $(\bmu_t)_{t =1}^L$, which in practice should be tuned to balance trend detection and noise reduction. To obtain the seasonal component we subtract the mean at each timestep, i.e. $\bx_{\text{season}} = (\bx_t - \bmu_t)_{t = 1}^L$, whereas the trend component is simply the moving average itself $\bx_{\text{trend}} = (\bmu_t)_{t = 1}^L$. Viewing both $\bx_{\text{season}}, \bx_{\text{trend}} \in \R^{L \times M}$ as matrices with $L$ time steps and $M$ channels, we apply the transformation:
\begin{align} \label{dlinear}
    \hat{\by} = \bW_{1} \bx_{\text{trend}} + \bW_{2}\bx_{\text{season}}
\end{align}
where $\bW_{1}, \bW_{2} \in \R^{O \times L}$ are learnable matrices and $\hat{\by} \in \R^{O \times M}$ is the forecast. Note that operations from Equations (\ref{moving_avg})$-$(\ref{dlinear}) are both linear, thus the transformation $\hat{\by} = f_{\theta}(\bx)$ is linear where $f_{\theta}$ is the network, making it a unique case among neural networks. 

\section{EMForecaster: Proposed DL-Empowered EMF Forecasting Architecture \label{sec:architecture}}
Given a univariate time series representing EMF exposure over time, we detail the pre-processing and the architecture of the proposed EMForecaster, shown in Figure~\ref{fig:architecture}. EMForecaster incorporates RevIN, hierarchical patching, and a mixing operation to process temporal patterns at multiple scales. By decomposing the input signal into localized patches, the architecture can efficiently capture both fine-grained temporal dynamics and long-range dependencies, enabling effective modeling compared to traditional approaches that process the entire sequence at once. In following subsections, we describe each component of the EMForecaster following the sequence of the pipeline shown in Figure~\ref{fig:architecture}.

\subsection{EMF Time Series Preprocessing}
Consider a time series $\bx = (x_t)_{t = 1}^T$, where $x_t \in \R$, $\forall t$. To eliminate outliers, we set a carefully selected threshold  $\delta >0$, where $x_t$ is replaced with a linear interpolation of its neighbors, $x_{t-1}$ and $x_{t+1}$. We then partition $(x_t)_{t = 1}^T$ into training, validation, and test datasets. We obtain the mean and standard deviation from the training set, and compute $z-$score to normalize the training, validation, and test datasets. Next, we apply the sliding window method with a stride of $1$, to obtain window pairs of size $L$ and $O$, representing the input and target, respectively. Thus, we obtain our dataset of $n$ input-output pairs, $\{(\bx_i, \by_i)\}_{i = 1}^n$ where $\bx_i \in \R^L$ is the lookback window (input), and $\by_i \in \R^O$ is its continuation (target), for any sequence $i$. 

\subsection{Reversible Instance Normalization (RevIN)}
We first apply RevIN to the input time series window $\bx$ before feeding it into any temporal modules \cite{revin}. RevIN has been proven to be effective for reducing the impact of distribution shifts and non-stationarity in time series  by using a symmetric structure to adjust and restore statistical information, leading to enhanced performance forecasting \cite{revin}. Given a window $\bx \in \R^L$, we first normalize by:
\begin{align} \label{revin}
    \bx^{(r)} = \text{RevIN}(\bx) = \gamma\Big(\frac{\bx \ominus \mu}{\sigma}\Big) \oplus {\delta},
\end{align}
where $\gamma, \delta \in \R$ are learnable affine parameters, and $\mu = \frac{1}{L} \sum_{i = 1}^L x_i$ and $\sigma = \sqrt{\frac{1}{L-1} \sum_{i = 1}^L (x_i - \mu)^2}$ are the mean and standard deviation, respectively of $\bx = (x_1, \dots, x_L)^T$. Here, $\oplus$ and $\ominus$ denote element-wise scalar-vector addition and subtraction, respectively, e.g., $\bv \oplus \alpha = (v_1 + \alpha, \cdots, v_k + \alpha)^T$ for any $\bv \in \R^k$ and $\alpha \in \R$, and similarly for $\ominus$. Once the \textit{candidate forecast}, $\hat{\by}_r$, is computed\textemdash the output of the spatiotemporal backbone defined in Section~\ref{sec:stb}\textemdash we perform the inverse operation (RevIN$^{-1}$) of Equation (\ref{revin}) by adding back the mean and multiplying the standard deviation into the channel output:
\begin{align}\label{revout}
    \hat{\by} = \text{RevIN}^{-1}(\hat{\by}^{(r)}) = \sigma\Big(\frac{\hat{\by}^{(r)} \ominus \delta}{\gamma} \Big) \oplus \mu,
\end{align}
where $\delta$ and $\gamma$ are the learnable affine parameters, and $\mu$ and $\sigma$ are the mean and standard deviation of $\bx$ from Equation (\ref{revin}). Note that:
$
    \text{RevIN} \circ \text{RevIN}^{-1} = \text{RevIN}^{-1} \circ \text{RevIN} = \mathbb{I},
$
where $\mathbb{I}$ is the identity function.

\subsection{Patching and Embedding}
Inspired from the PatchTST and Vision Transformer (ViT) methods \cite{vit, patchtst}, we consider a patching mechanism along with a transformer encoder backbone. Given the demonstrated effectiveness of patching in time series forecasting compared to non-patching methods \cite{logtrans, pits}, we customize this mechanism for our architecture. 
After passing the input time series $\bx$ to obtain $\bx^{(r)} = \text{RevIN}(\bx)$, we apply a patching module to split up the sequence into equally-sized contiguous subsequences. Given a pre-determined patch dimension $P$, and patch stride $S$\textemdash similar to a convolutional kernel\textemdash we obtain $N = \lfloor \frac{L-P}{S}\rfloor + 1$ patches in total:
\begin{align} \label{patching}
    \bx^{(p)} = \text{Patcher}(\bx^{(r)}),
\end{align}
where $\bx^{(p)} \in \R^{N \times P}$. Before applying patching, we pad $\bx^{(r)}$ with zeros at the end of the time series to ensure there are enough positions to extract all $N$ complete patches of size $P$. After patching, we perform patch embedding defined by:
\begin{align}
    \bx^{(d)} &= \text{PatchEmbed}(\bx^{(p)}) \\ 
    &= [\bW_d(\bx^{(p)}_1) \   \bW_d (\bx^{(p)}_2)  \cdots   \bW_d(\bx^{(p)}_N)]^T \\
    &= \bx^{(p)} \bW_d^T,
\end{align}
where $\bx^{(p)}_i \in \R^P$ is the $i^{\mathrm{th}}$ row of $\bx^{(p)}$, $\bx^{(d)} \in \R^{N \times D}$ is the resultant embedding, and $\bW_d \in \R^{D \times P}$ is a learnable weight matrix which embeds each patch of size $P$ to a (typically larger) dimension of size $D$, which we refer to as the patch embedding dimension. This is in contrast to several other embedding mechanisms found in other DL time series models, which may apply an embedding along the sequence dimension with $\bx^{(d)} = \bW_d \bx^{(p)}$ with $\bW_d \in \R^{D \times N}$. The patch-wise embedding approach learns local temporal patterns while maintaining parameter efficiency, as the embedding weights are shared across all patches and independent of the sequence length. Selecting an appropriate patch dimension $P$ and patch embedding dimension $D$ for the task is critical, as $P$ controls the temporal receptive field of local pattern learning, while $D$ determines the richness of the learned representations and the model's capacity to capture complex temporal behavior.


\subsection{Spatiotemporal Backbone (STB) \label{sec:stb}}
MLPMixer, a model proposed initially for computer vision, processes information through two distinct MLP operations: one that mixes information across tokens and another that mixes features within each token \cite{mlpmixer}. MLPMixer demonstrated that MLP-based architectures could achieve state-of-the-art performance in computer vision, even when compared to several modern Transformer variants \cite{vit}, challenging the fact that convolutions or self-attention mechanisms are necessary for visual processing. TSMixer adapts the MLPMixer architecture for time series by utilizing two distinct MLP operations per block: one that mixes information across temporal locations and another that mixes features within each time step. For a layer $l$, the temporal MLP $f_{\theta^{(l)}}: \R^{N\times D} \to \R^{N\times D}$ is defined as:
\begin{align} \label{temporal_mlp}
    f_{\theta^{(l)}}(\bu) =  \bW_{\theta_2^{(l)}}\big(\sigma(\bW_{\theta_1^{(l)}}\bu)\big),
\end{align}
for any $\bu \in \R^{N \times D}$, where $\bW_{\theta_1^{(l)}} \in \R^{D_h \times N}$ and $\bW_{\theta_2^{(l)}} \in \R^{N \times D_h}$ are learnable matrices with parameters $\theta_1^{(l)}$ and $\theta_2^{(l)}$ respectively, $\sigma$ is a nonlinearity (e.g., ReLU), and $D_h$ is the hidden dimension of the MLP. Similarly, the patch MLP $g_{\phi^{(l)}}: \R^{N \times D} \to \R^{N \times D}$ is defined as:
\begin{align} \label{channel_mlp}
    g_{\phi^{(l)}}(\bu) = \Big(\bW_{\phi_2^{(l)}}(\sigma(\bW_{\phi_1^{(l)}}\bu^T)\Big)^T,
\end{align}
for any $\bx \in \R^{N \times D}$, where $\bW_{\phi^{(l)}} \in \R^{D_h \times D}$ and $\bW_{\phi_2^{(l)}} \in \R^{D \times D_h}$ are learnable matrices with parameters $\phi_1^{(l)}$ and $\phi_2^{(l)}$, respectively. The hidden dimension $D_h$ and nonlinearity $\sigma$ are the same as in the temporal MLP.

We provide the patch-embedded time series $\bx^{(d)} \in \R^{N \times D}$ as an input to STB composed of $K$ blocks:
\begin{align}
    \bx^{(b)} = \text{STB}(\bx^{(d)}) = \left(\text{Block}^{(K)} \circ \cdots \circ \text{Block}^{(1)}\right)(\bx^{(d)}).
\end{align}
where the $l^{\mathrm{th}}$ block is modeled as:
\begin{align}
    \text{Block}^{(l)}(\bu) = \bu + f_{\theta^{(l)}}(\bu) + g_{\phi^{(l)}}\big(\bu + f_{\theta^{(l)}}(\bu)\big).
\end{align}
The above is the explicit form of composing $f_{\theta^{(l)}}$ and $g_{\phi^{(l)}}$ with two residual connections \cite{resnet}, which can be written equivalently as follows:
\begin{align} \label{tsmixer}
    \bu' &= \bu + f_{\theta^{(l)}}(\bu), \\
    \text{Block}^{(l)}(\bx) &= \bu' + g_{\phi^{(l)}}(\bu').
\end{align}
where, $f_{\theta^{(l)}}$ ``mixes" the temporal dimension $N$ and $g_{\phi^{(l)}}$ ``mixes" the embedding dimension $D$.

Applying the STB to patch-embedded data represents a novel fusion where the mixer operates on learned patch representations rather than raw temporal slices, enabling the capture of hierarchical patterns at multiple scales. The combination of patch embeddings and mixing operations allows the model to learn relationships between these higher-level temporal abstractions while maintaining the architectural simplicity of MLP-based mixing.
We then pass the output of the STB through a nonlinear activation $\sigma$ and a layer normalization module \cite{layernorm}, i.e.,
$
    \bx^{(a)} = \text{LayerNorm}(\sigma(\bx^{(b)})),
$
and flatten $\bx^{(a)}$ to
$
    \bx^{(f)} = \text{Flatten}(\bx^{(a)}),
$
where $\bx^{(f)} \in \R^{({N  D)}}$, in which we obtain the candidate forecast by:
\begin{align}
    \hat{\by}^{(r)} = \text{Head}(\bx^{(f)})= \bW_{\text{head}} \bx^{(f)},  \\ \text{where} \quad \bx^{(f)} = \text{Flatten}(\text{LayerNorm}(\sigma(\bx^{(b)}))),
\end{align}
where $\bW_\text{head} \in \R^{O \times (N  D)}$ is a learnable weight matrix, and $\hat{\by}^{(r)} \in \R^{O}$ is the candidate forecast, which is  refined to obtain the final forecast by $\hat{\by} = \text{RevIN}^{-1}(\hat{\by}^{(r)})$ given by  (\ref{revout}).

\section{EMForecaster with Conformal Prediction \label{sec:conformal}}
CP is a powerful framework for uncertainty quantification  that provides rigorous statistical guarantees \cite{cp}. The goal of CP is to provide a \emph{prediction region}, $\Gamma^\alpha$, for a given \textit{significance level \text{(or error} rate}) $\alpha \in (0,1)$, ensuring that the true outcome falls within this region with a probability of at least $(1 - \alpha)$. Most CP approaches rely on an \emph{inductive approach}, where predictions are generated using a combination of an \emph{underlying model} $f$ and an additional \emph{calibration set}—a technique referred to as \emph{inductive conformal prediction (ICP)} \cite{angelopoulos2020uncertainty,papadopoulos2008inductive}. Rather than providing point predictions, such as a single numerical value in regression, models calibrated with ICP generate a continuous interval in which ground truth is theoretically guaranteed to be contained with probability $1-\alpha$. 

In this section, we first describe CP in the context of regression with functions of the form $f: \R^L \to \R$. We then discuss the extension of CP to regression functions of the form $f: \R^L \to \R^O$  to predict multiple future time points simultaneously. Next, we discuss the evaluation metrics and the proposed metric referred to as Trade-off Score (TOS).

\subsection{CP: Single Time-Step Forecast}
While CP has been widely studied, its application to time series forecasting was limited due to the \emph{exchangeability} condition which requires that any reordering of the dataset is equiprobable. 
\begin{definition}[Exchangeability]
Given a dataset with $n$ observations $ \mD = \{(\bx^{(i)}, y^{(i)})\}_{i=1}^n \sube \R^{T} \times \R$, we say that it is \textit{exchangeable} if any of its $n!$ permutations are equiprobable. Note that independent and identically distributed (i.i.d.) observations satisfy exchangeability.
\end{definition}
Given the exchangeability, the trustworthiness or reliability of a conformal predictor is described by the conformal coverage guarantee  defined below.
\begin{property}[Conformal coverage guarantee] \label{coverage_guarantee}
Under the exchangeability assumption, any conformal predictor will return the prediction region $\Gamma^\alpha(\mathbf{x}^{(i)})$ such that the probability of error $y^{(l+1)} \notin \Gamma^\alpha(\mathbf{x}^{(l+1)})$ is not greater than $\alpha$, that is:
\begin{align}
    \mathbb{P}[y^{({l+1})} \in \Gamma^\alpha(\mathbf{x}^{(l+1)}) \mid \mathcal{D}] \geq 1 - \alpha.
\end{align}
\end{property}
Note that the exchangeability assumptions and validity properties are \emph{distribution-free}; that is, there are no required distributional assumptions on the underlying data $\mathcal{D}$, and applies to any underlying predictive model so long as the exchangeability assumption is satisfied on $\mD$.

Time series data, however, exhibits temporal dependencies, which inherently violating the exchangeability condition in most scenarios. As a result, directly applying CP to forecast intervals in time series data may not provide theoretical guarantees. Recent advancements have demonstrated that the exchangeability assumption can be relaxed through problem reframing this to the exchangeability of windows, allowing CP to be effectively applied to time series tasks \cite{enbpi, ctsf}. 

The inductive variant of CP divides the dataset into two subsets: a proper training set and \emph{calibration set}, such that $\mathcal{D} = \mathcal{D}_{\text{train}} \cup \mathcal{D}_{\text{cal}}$. Denote $n = |\mD|$ and $m = |\mD_{\text{cal}}|$. The training set is used to fit the underlying model $f$, while the calibration set is utilized to compute the \emph{nonconformity scores} $\eta(\mathcal{D}, (\bx^{(i)}, y^{(i)}))$ for all $(\bx^{(i)}, y^{(i)}) \in \mD_{\text{cal}}$, which measures the distribution errors on the underlying model. CP guarantees validity for any choice of nonconformity score, even random ones, but in regression settings, a commonly used score is defined as:
\begin{align}
    R_i = \eta(\mathcal{D}, (\mathbf{x}^{(i)}, y^{(i)})) = d\big(f(\mathbf{x}^{(i)} \mid \mathcal{D}), y^{(i)}\big),
\end{align}
where $d: \mathbb{R} \times \mathbb{R} \to \mathbb{R}$ is a distance metric. While the choice of $d$ is flexible, it should reflect the problem's objectives with respect to the dataset and task. The residuals $\{R_i\}_{i=1}^m$, computed from the calibration set, form an empirical distribution. The \emph{critical nonconformity score}, $\hat{\epsilon}$, is selected as the $(1-\alpha)$-quantile of this distribution. To account for finite-sample effects, a correction is applied by considering the $\lceil (m+1)(1-\alpha) \rceil$-th smallest residual. For any new input $\mathbf{x}^{(n+1)}$, the prediction interval is defined by:
\begin{align} \label{gamma_interval}
\Gamma^\alpha(\mathbf{x}^{(n+1)}) = [\hat{y}^{(n+1)} - \hat{\epsilon}, \hat{y}^{(n+1)} + \hat{\epsilon}],
\end{align}
where $\hat{y}^{(n+1)} = f(\mathbf{x}^{(n+1)})$ is the model prediction.

\begin{theorem}[Vovk, Gammerman, and Saunders \cite{vovk1999machine}] \label{coverage_guarantee_thm}
    Suppose $\Gamma^\alpha$ is defined as in Equation (\ref{gamma_interval}), then $\Gamma^\alpha$ is a conformal predictor and Property (\ref{coverage_guarantee}) is satisfied.
\end{theorem}

\subsection{CP:  Multiple Time-Step Forecast \label{sec:ctsf}}
In univariate time series forecasting with horizon $O > 1$ and underlying model $f: \R^L \to \R^O$, our goal is to generate prediction intervals for each future time point. Specifically, for a model that produces predictions $f(\mathbf{x}^{(n+1)}) = \hat{\by} = (\hat{y}_1, \dots, \hat{y}_O)^T$, we must construct intervals as follows: $$[\hat{y}_t^{(n+1)} - \hat{\vep}_t, \:\: \hat{y}_t^{(n+1)} + \hat{\vep}_t], \:\: 1 \leq t \leq O.$$
For any new observation $(\mathbf{x}^{(n+1)}, \mathbf{y}^{(n+1)})$ where $\mathbf{y}^{(n+1)} = (y_1^{(n+1)}, \dots, y_O^{(n+1)})^T$, and any time step $1 \leq t \leq O$, these intervals should satisfy the following coverage guarantee similar to Property (\ref{coverage_guarantee}):
\begin{align} \label{ts_coverage_guarantee}
\mathbb{P}\big[y_t^{(n+1)} \in [\hat{y}_t^{(n+1)} - \hat{\vep}_t, \hat{y}_t^{(n+1)} + \hat{\vep}_t ]\big] \geq 1 - \alpha
\end{align}
For any example $(\bx^{(i)}, \by^{(i)})$, the nonconformity score can be generalized to:
\begin{align}
    R_i = \begin{bmatrix}
        |y_{1}^{(i)} - \hat{y}_1^{(i)}|, \dots, |y_{O}^{(i)} - \hat{y}_O^{(i)}|
    \end{bmatrix}
\end{align}
In \cite{ctsf}, the authors indicate that the only restriction needed is that the underlying model must map to all $O$ predictions simultaneously, rather than recursively, such as in Bayesian forecasting methods. Given this constraint, as all $O$ predicted values are obtained from the same representation, we employ a Bonferroni correction when obtaining the critical nonconformity scores, which are obtained by considering the $\lceil (m+1)(1-\alpha/O) \rceil$-th smallest residuals $\hat{\vep}_1, \dots, \hat{\vep}_O$ from each score distribution. By defining:
\begin{align} \label{ts_intervals}
    \Gamma^{\alpha}_t(\hat{y}_t) = [\hat{y}_t - \hat{\vep}_t, \hat{y}_t + \hat{\vep}_t]
\end{align}
for all $1 \leq t \leq O$, we obtain our valid conformal predictor.

\begin{theorem} 
    Let $\mD = \{(\bx^{(i)}, \by^{(i)})\}_{i = 1}^n$ be the dataset of exchangeable time series windows, where $\by^{(i)} \in \R^O$ is the continuation of time series window $\bx^{(i)} \in \R^L$, for every $1 \leq i \leq n$, generated from the same underlying distribution. Let $f: \R^L \to \R^O$ be a forecasting model which maps to all forecasted values simulateneously. Then for any $\alpha \in (0,1)$, using the method in Equation (\ref{ts_intervals}) we have that:
    \begin{align}
        \mathbb{P}[y_{t} \in \Gamma^{\alpha}_t(\hat{y}_t), \textnormal{ for } 1 \leq t \leq O] \geq 1 - \alpha
    \end{align}
\end{theorem}

\subsection{Evaluation Metrics}
For a set of predictions $\hat{\by}_1, \dots, \hat{\by}_n$ with corresponding target sequences $\by_1, \dots, \by_n$, the \textit{independent coverage} (IC) and \textit{joint coverage} (JC) are defined by:
\begin{align}
   \text{IC} &= \frac{1}{nO} \sum_{i=1}^n \sum_{t = 1}^O \mathbbm{1}[y_t^{(i)} \in \Gamma_t^{\alpha}(\hat{y}_t^{(i)})] \\
   \text{JC} &= \frac{1}{n} \sum_{i=1}^n \mathbbm{1}\left\{\bigwedge_{t=1}^O y_t^{(i)} \in \Gamma_t^{\alpha}(\hat{y}_t^{(i)})\right\}
\end{align}
where $\mathbbm{1}$ is the indicator function and $\bigwedge$ denotes the logical \texttt{AND} operation.  $\text{IC}$ measures the proportion of examples where the true value falls within the prediction interval at each horizon time $t$ separately, while JC measures the proportion of examples where the true values fall within their respective prediction intervals across the full horizon time points simultaneously. While the empirical IC and JC typically increase with lower significance levels $\alpha$, the \textit{mean interval width} (MIW) of the prediction given below:
\begin{align}
\text{MIW} = \frac{1}{O}\sum_{t = 1}^O 2\hat{\vep}_t,
\end{align}
typically increases as well, which provides the average measure of uncertainty for model predictions. Therefore, a fundamental trade-off exists between coverage and interval width: models achieving higher JC often do so at the cost of wider prediction intervals. 

\textit{The core challenge is therefore to minimize the MIW while maximizing JC and IC across a chosen significance level $\alpha$. }

\subsection{The Trade-off Score (TOS) \label{sec:tos}}
Given the fundamental trade-off between achieving optimal coverage while maintaining efficient prediction intervals, we propose a unified scoring metric that synthesizes JC, IC, and MIW into a single measure bounded in [0,1]
referred to as weighted-average coverage (WAC) defined below:
\begin{align} \label{wac}
\text{WAC} = \frac{\beta \text{JC} + (1 - \beta) \text{IC}}{2}
\end{align}
where $\beta \in [0,1]$ is a hyperparameter controlling the relative importance of joint coverage versus independent coverage. While the choice of $\beta$ should reflect the practitioner's objectives, we recommend $\beta > \frac{1}{2}$ as joint coverage typically presents a more important and stricter criterion in CP.
To incorporate the width of prediction regions, consider a collection of $k$ conformal predictors derived from $k$ models with corresponding MIWs $m_1, \dots, m_k \in \mathbb{R}^+$. We standardize these measures using $z$-score normalization $
z_i = \frac{\mu - m_i}{\sigma},
$
where $\mu$ and $\sigma$ denote the sample mean and standard deviation of $\{m_1, \dots, m_k\}$, respectively. Finally, we introduce our proposed trade-off score (TOS), a  metric that  combines both coverage validity and prediction width. For each conformal predictor $i$, we define:
\begin{align} \label{tos}
\text{TOS}(i) = \lambda \text{WAC} + (1-\lambda) \frac{1}{1+e^{z_i}}
\end{align}
where $\lambda \in [0,1]$ controls the balance between WAC and normalized MIW. The reflected sigmoid transformation of the $z$-scores ensures the second term remains bounded in [0,1], resulting in $\text{TOS}(i) \in [0,1]$ for all conformal predictors. 
This unified metric acknowledges that an ideal conformal predictor should not only achieve the desired coverage levels, but do so with reasonably tight prediction intervals. Without such a combined metric, we might favor methods that achieve perfect coverage at the cost of excessively wide intervals, or methods that produce deceivingly narrow intervals but fail to maintain reliable coverage. The TOS enables practitioners to make these trade-offs explicit through the tuning parameters $\beta$ and $\lambda$, while ensuring fair comparison across different methods through appropriate normalization. 
 For meaningful comparisons, it is recommended to evaluate against a diverse set of  methods that span different architectural families and methodological approaches. This helps ensure robust normalization and provides comprehensive performance context.

\section{Experimental Set-up and Data Analysis \label{sec:experiments}}
In this section, we describe each of the selected datasets along with our experimental configuration for point forecasting and conformal forecasting. Our benchmarks include both \textit{short-term} and \textit{long-term} EMF exposure series. We define short-term to be any range within 24 hours, and long-term to exceed 24 hours. We analyze the characteristics of EMF exposure in several datasets through spectral decomposition, stationarity testing, and spatial correlation analysis.
\subsection{EMF Datasets}
\subsubsection{Long-term EMF Series (Italy)}

Adda \textit{et al.} \cite{adda2023highnoon} presented long-term EMF  measurements from four locations within  Rome and Turin, Italy. Table~\ref{table:italy} describes all measurement sites and their respective dataset specifications such as the year of recording, total length of each time series (i.e. duration of measurement), and reported sampling rate $(\Delta t)$. Measurement sites include the Department of Electronic Engineering (DEE) at the University of Rome Tor Vergata, a University Hospital (UH) in Rome, Polytechnic University of Turin (PT), and the Porta Nuova Train Station (TS). 

\begin{figure}[ht]
    \centering
    \includegraphics[width=\columnwidth]{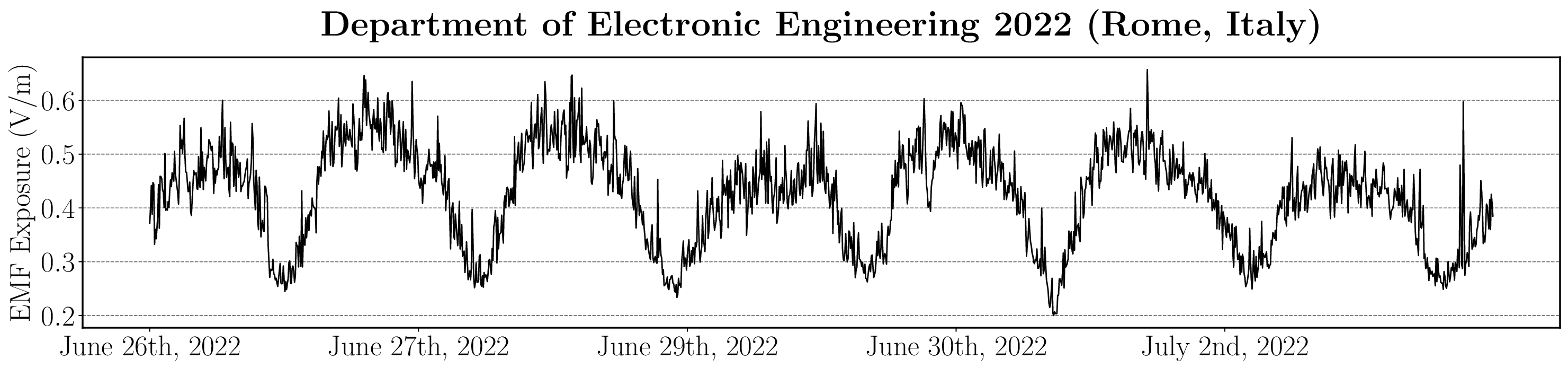} \\
    
    
    \includegraphics[width=\columnwidth]{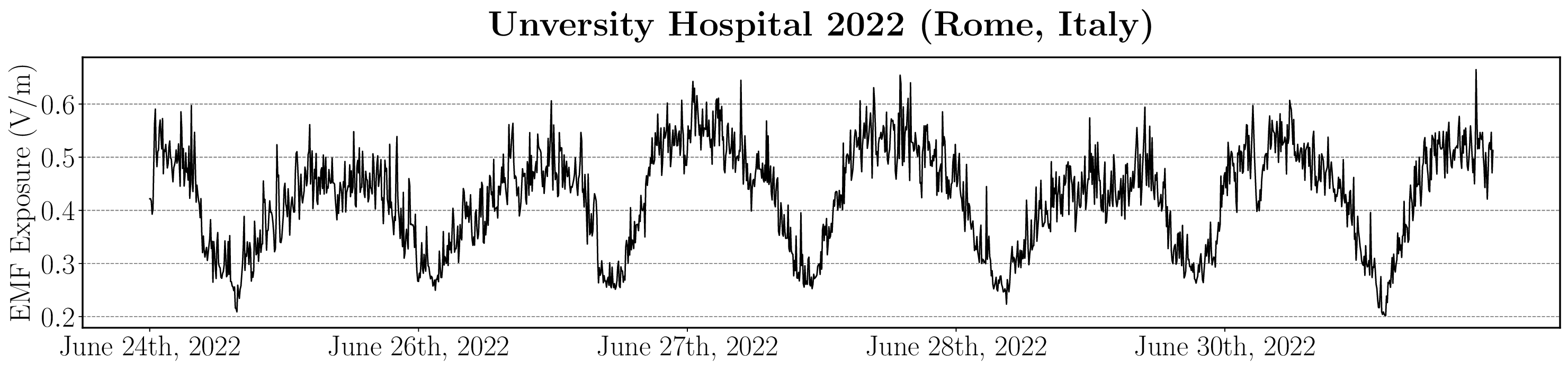} \\
    
    \includegraphics[width=\columnwidth]{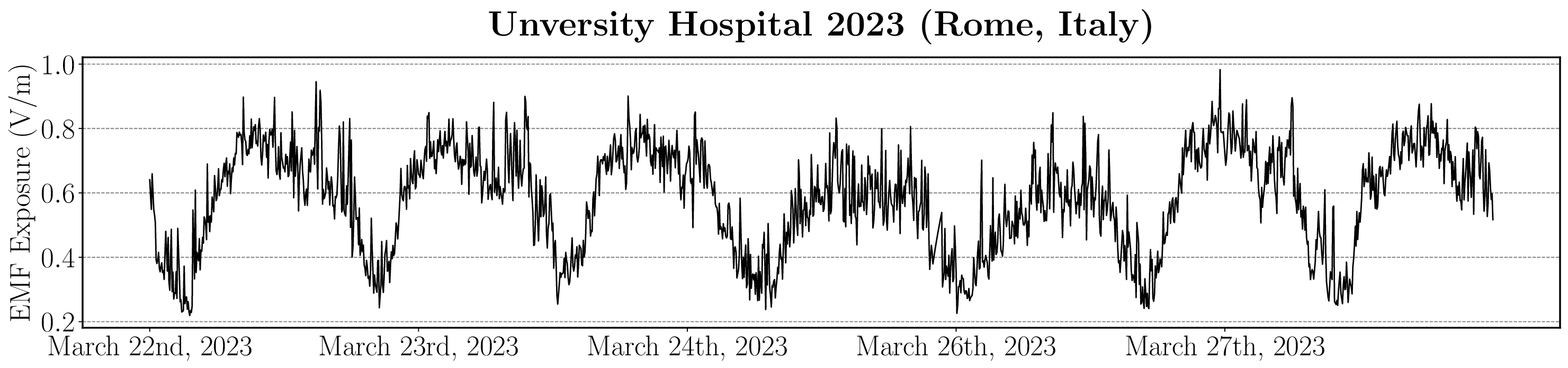} \\
    
    \includegraphics[width=\columnwidth]{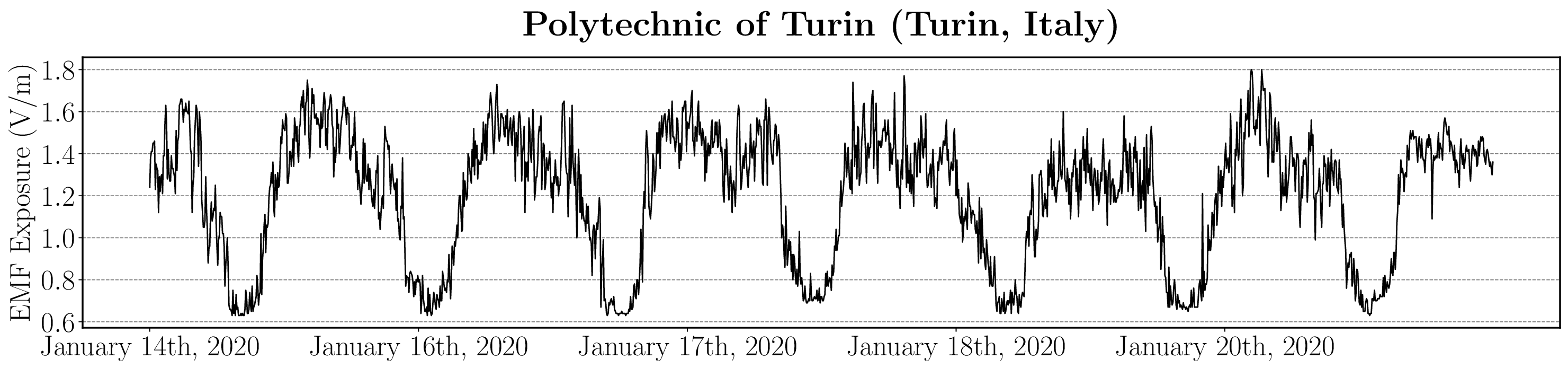} \\
    
    \includegraphics[width=\columnwidth]{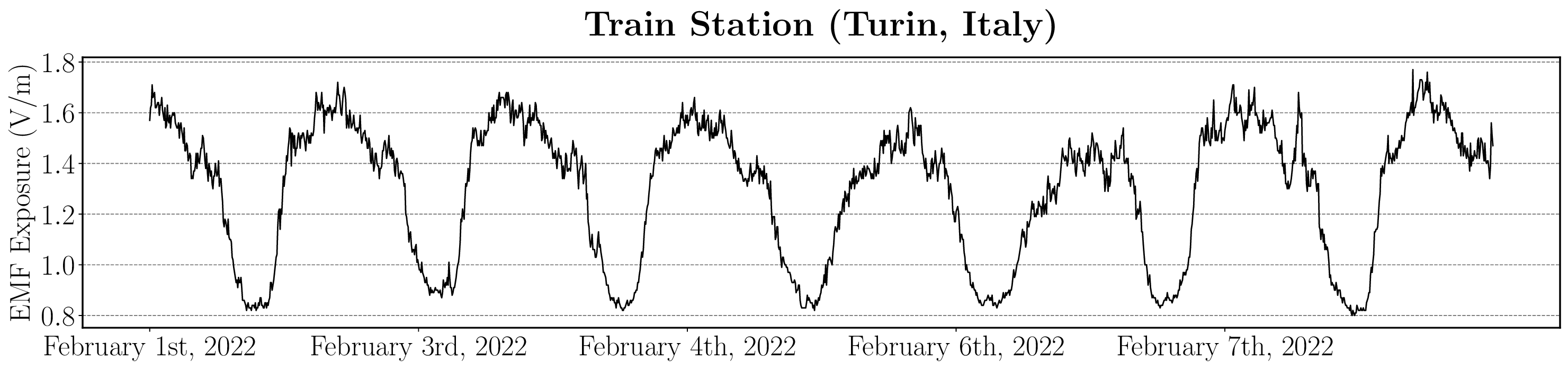}

    \includegraphics[width=\columnwidth]{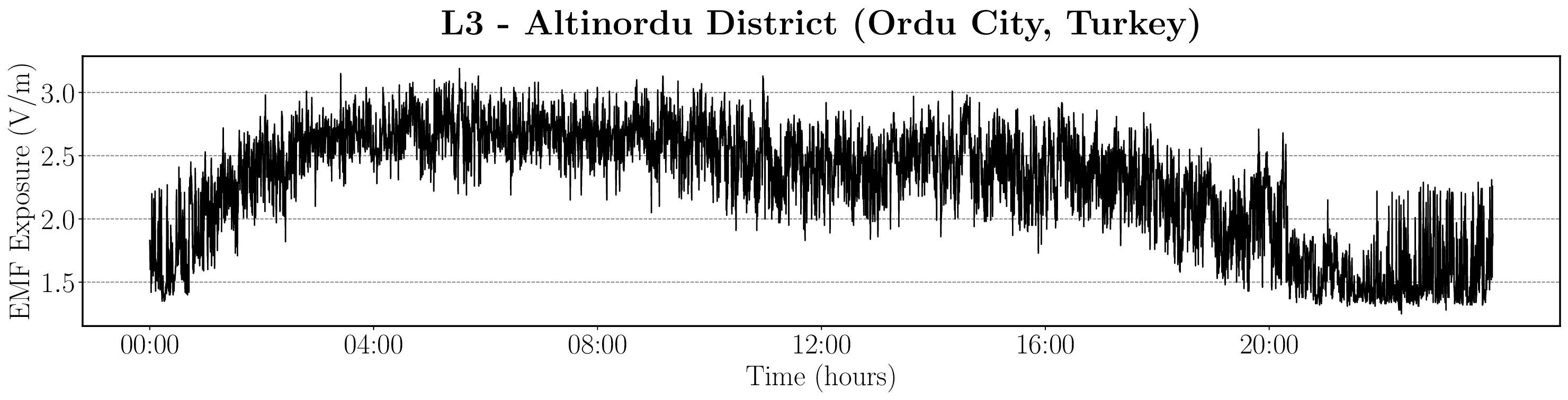} \\
    
    \includegraphics[width=\columnwidth]{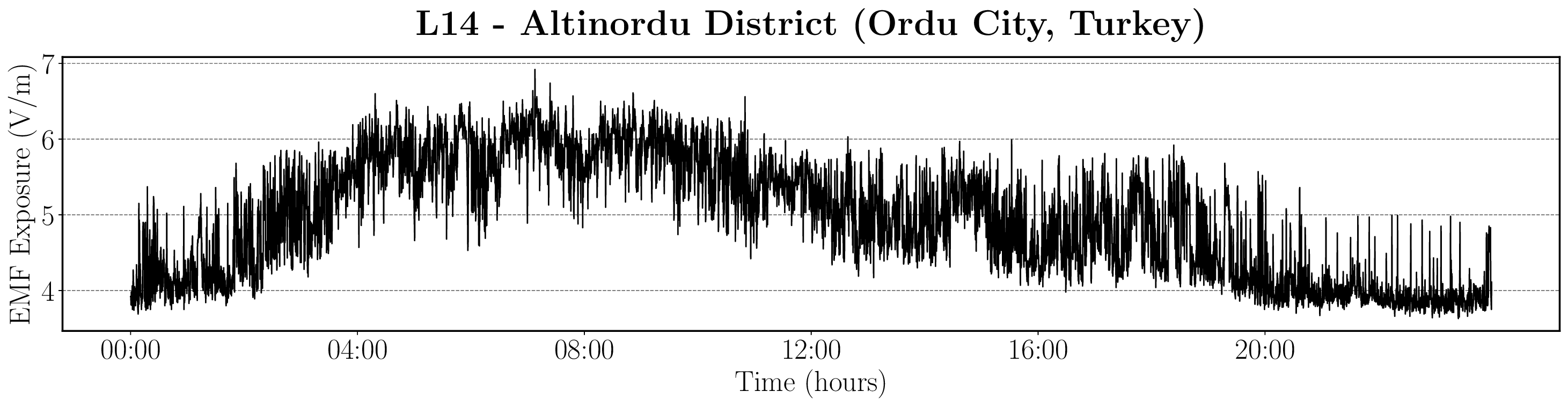} 

    \caption{Visualization of the EMF exposure (V/m) over time from five locations in the Italy (DEE 22', UH 22', UH 23', PT, TS) and two locations from the Turkey dataset (L3, L14).}
    \label{fig:timeseries}
\end{figure}

\begin{table}[t]
\centering
\scriptsize
\caption{Long-term EMF datasets from Italy with reported the duration of measurement, year of recording, and reported sampling rate $(\Delta t)$.}
\begin{tabular}{ccccc}
\toprule
\textbf{City}   & \textbf{Location} & \textbf{Duration} & \textbf{Year} & \boldmath$\Delta t$ \\
\midrule
Rome   & Dept. of Electronic Engineering (DEE 22')  & 190 days    & 2022 & 6 min \\
Rome   & Dept. of Electronic Engineering (DEE 23')  & 491 days    & 2023 & 6 min \\
Rome   & University Hospital (UH 22')              & 139  days   & 2022 & 6 min \\
Rome   & University Hospital (UH 23')              & 592  days   & 2023 & 6 min \\
Turin  & Polytechnic University of Turin (PT)            & 168 days    & 2020 & 6 min \\
Turin  & Train Station (TS)             &  29 days & 2022 & 6 min \\
\bottomrule
\label{table:italy}
\end{tabular}
\end{table}
Although each location exhibits different characteristics in their EMF distribution, each site displays similar daily exposure patterns with cyclical behaviour, as shown in Figure~\ref{fig:timeseries}. Note that $\Delta t$ represents the 6 min sampling interval (pre-processed from the data provider), where the true sampling rate of the device is 3~sec. For each location within Rome, narrow-band monitoring was performed using an Anritsu MS27102A spectrum monitor (9 kHz – 6 GHz) connected to a Keysight N6850A omnidirectional antenna (20 MHz – 6 GHz) and the required EMF aggregated over all frequencies is then obtained by computing the root sum squared values of the EMF measured at narrow-band frequencies. In Turin, a Narda 8059 wideband monitor with an electric field sensor (100 kHz – 7 GHz) was used. For more information regarding each dataset, we refer the readers to \cite{adda2023highnoon}.

\begin{figure}[ht]
    \centering
    \includegraphics[width=\columnwidth]{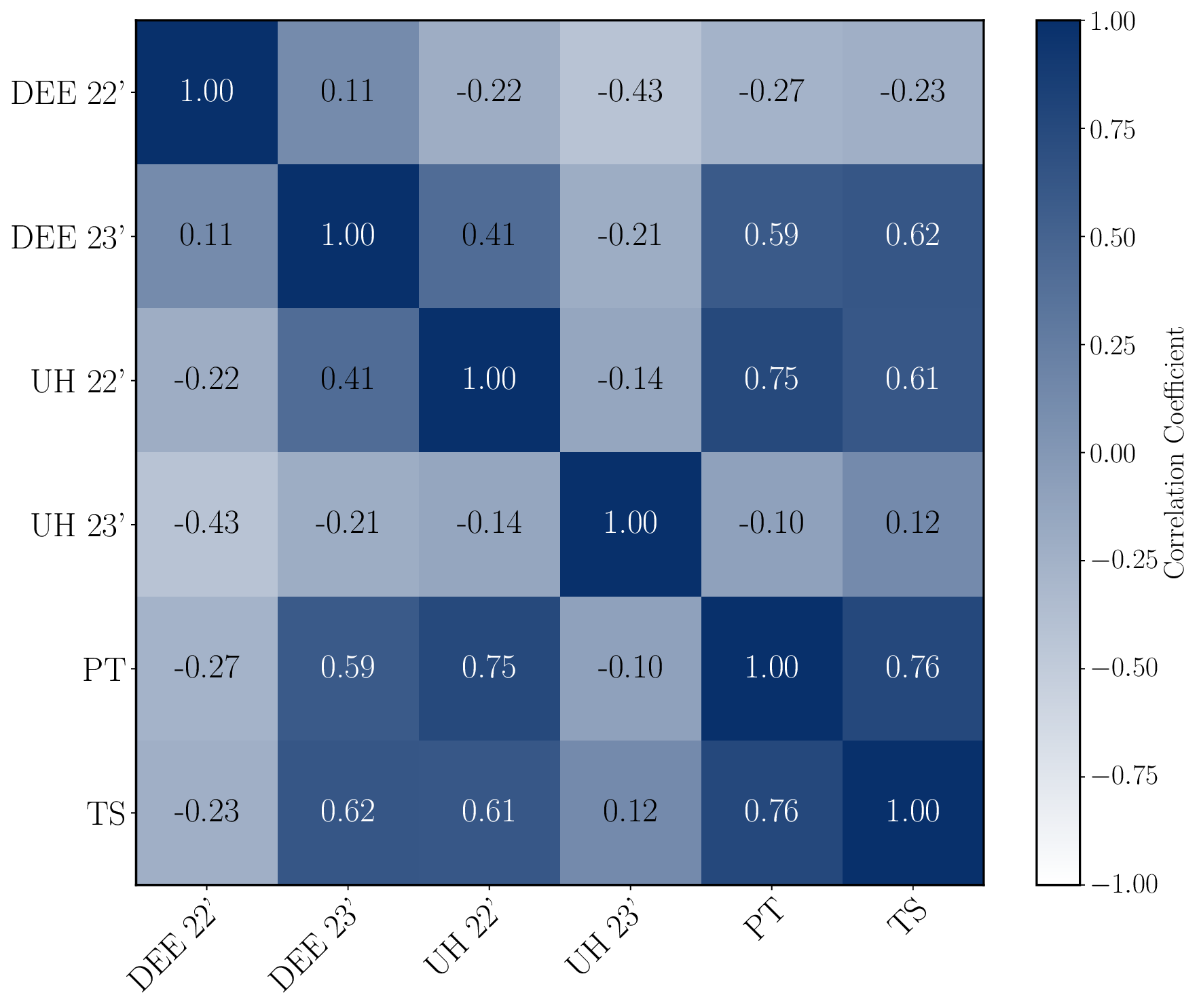} \\
    \caption{Correlation heatmap of EMF exposure over a one week period for all 4 locations in Italy.}
    \label{fig:italy_heatmap}
\end{figure}

\begin{figure}[ht]
    \centering
    \includegraphics[width=\columnwidth]{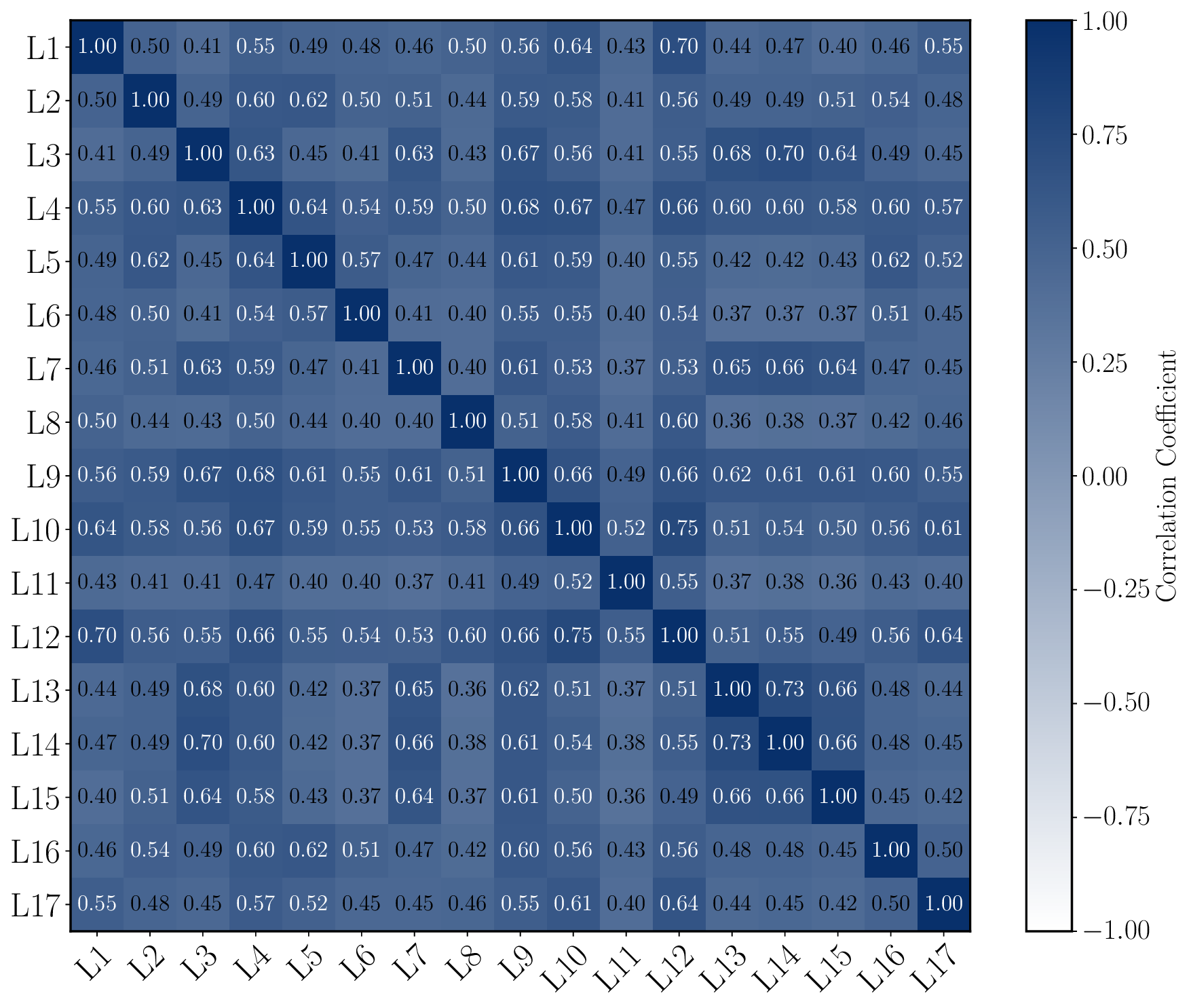} \\
    \caption{Correlation heatmap of EMF over a 24 hour (total) period for all 17 locations within the dataset from Turkey.}
    \label{fig:turkey_heatmap}
\end{figure}

\begin{figure}[ht]
    \centering
    \begin{subfigure}{0.45\textwidth}
        \centering
        \includegraphics[totalheight=2in, width=\textwidth]{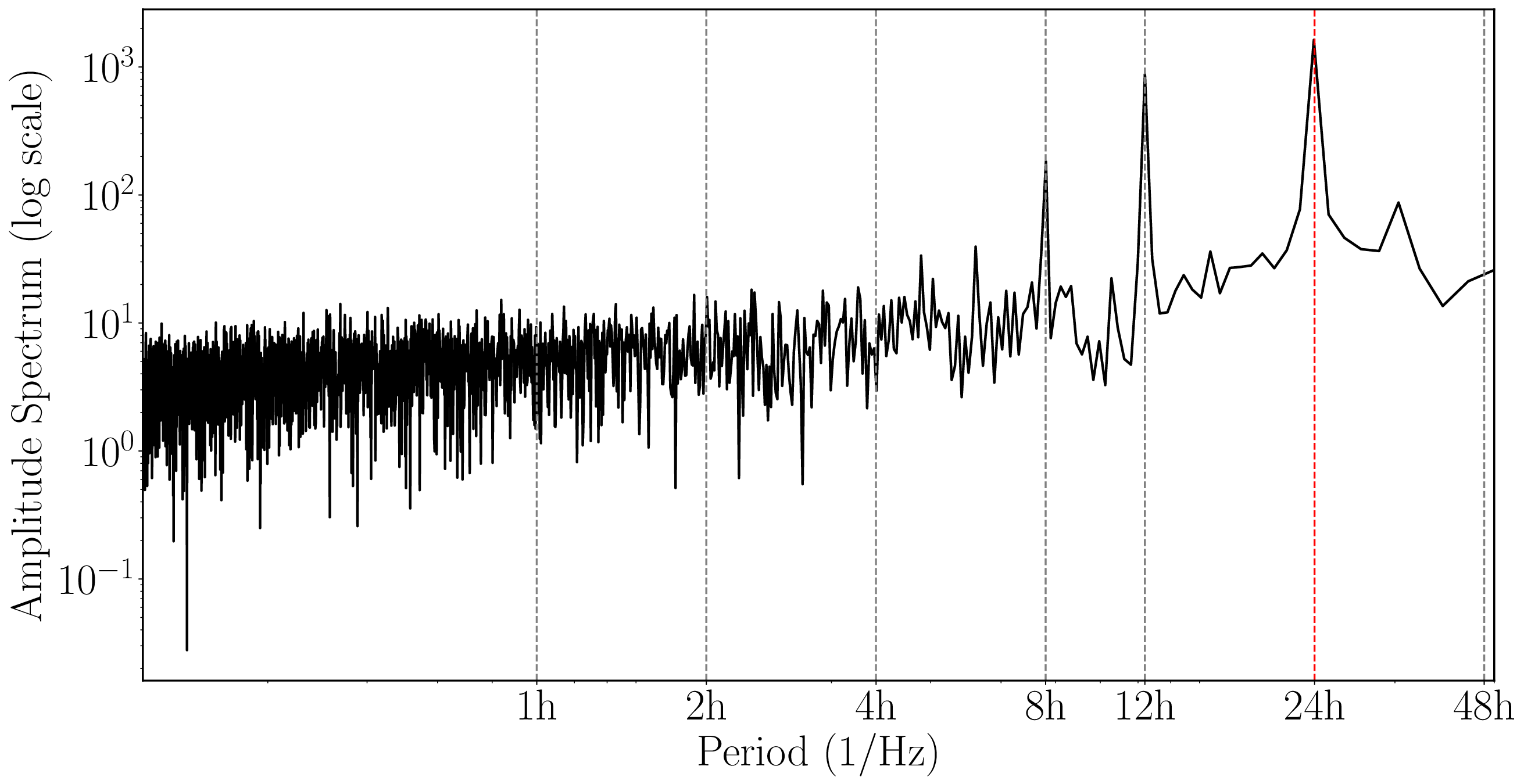}
        \caption{FFT magnitude (DEE 22') with respect to the period (1/Hz).}
        \label{fig:dee22_fft}
    \end{subfigure}
    \hfill
    \begin{subfigure}{0.45\textwidth}
        \centering
        \includegraphics[totalheight=2in, width=\textwidth]{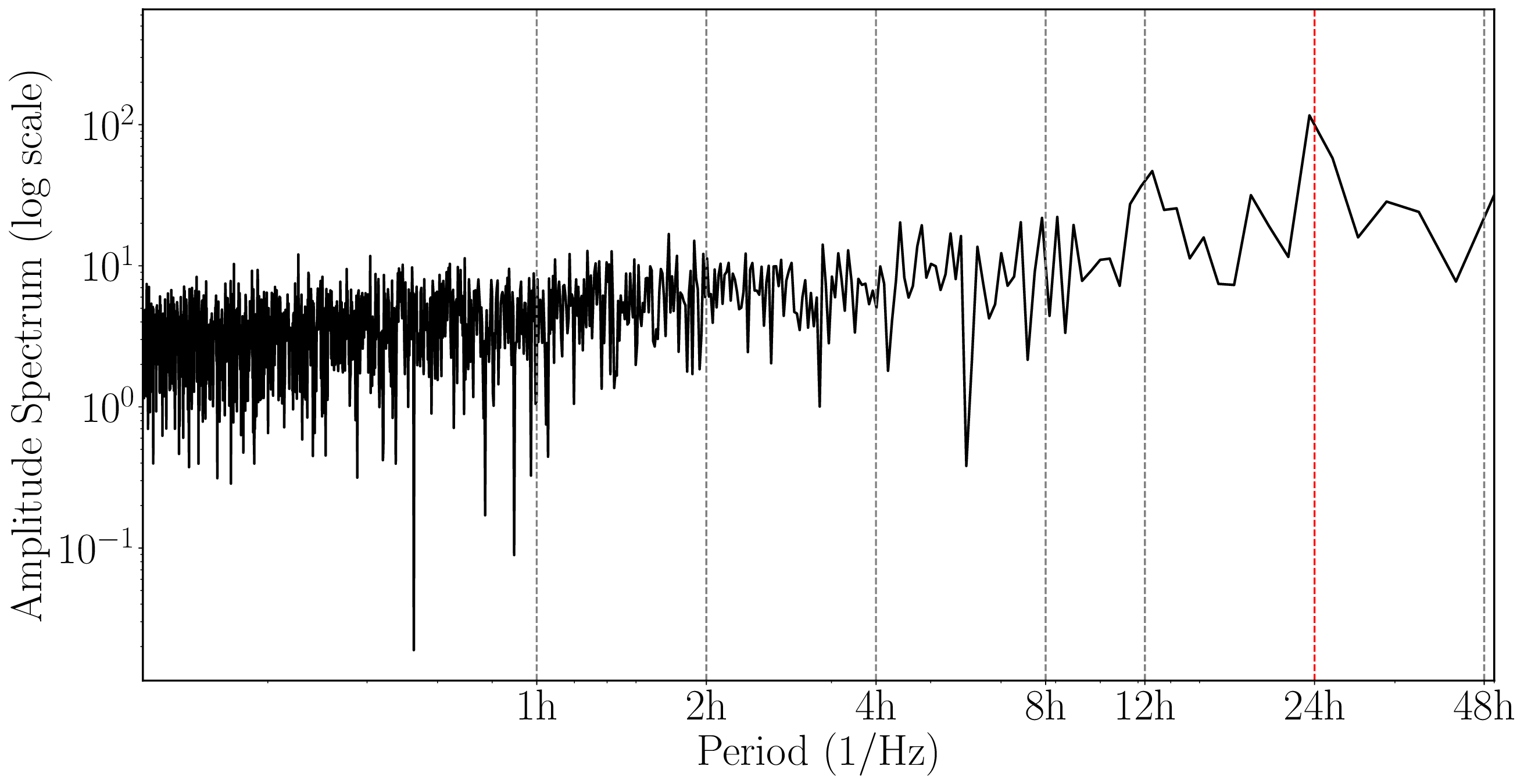}
        \caption{FFT magnitude (UH 22') with respect to the period (1/Hz).}
        \label{fig:uh22_fft}
    \end{subfigure}
    \caption{FFT magnitude plots of DEE 22' and UH 22'.}
    \label{fig:fft_plots}
\end{figure}



\subsubsection{Short-term EMF Series (Turkey)} 
In \cite{kurnaz2020rfemf}, Kurnaz \textit{et al.} collected EMF exposure data in the Altınordu District of Ordu City, Turkey. Short-term ($\leq 24$hr) broadband EMF measurements were conducted at 17 locations along the main streets of Altınordu, covering a frequency range of 100 kHz to 3 GHz, with a sampling rate of 15 seconds and duration of 24 hours. The authors used a PMM-8053 EMF meter equipped with an EP-330 electric field isotropic probe for the wideband measurements.

 


\subsection{Data Analysis \& Visualization}
\subsubsection{Long-Term vs. Short-Term Forecasting}
The datasets from Italy and Turkey exhibit fundamental differences in their exposure measurements that significantly impact our analysis and potential model forecasting capabilities. The datasets from Italy contain extensive recordings spanning up to 592 days across six different locations, showing clear and consistent daily cyclical patterns throughout the extended monitoring period. On the other hand, the dataset from Turkey covers 24-hours periods for each of the 17 locations, {but comes with more frequent measurements at 15-sec intervals compared to Italy's 6 min reported sampling rate.} These contrasting characteristics indicate that our forecasting approach for the dataset from Italy can effectively model both the prominent daily cycles and their deviations, whereas the dataset from Turkey presents us with a more challenging prediction scenario due to its shorter observation window and less apparent periodicity, despite its higher temporal resolution. Our decision to maintain the 15-sec sampling rate for the dataset from Turkey rather than downsampling to match the 6 min intervals of datasets from Italy was driven by data quantity considerations. Averaging to 6 min intervals would have drastically reduced our already limited amount of data available from the 24-hours recordings, which would be insufficient for a fair comparison between our proposed DL model and baselines.

\subsubsection{Correlation of Measurement Sites}
Figures~\ref{fig:italy_heatmap} and \ref{fig:turkey_heatmap} illustrate the correlation between measurement sites for Italy and Turkey through heatmap visualizations, respectively. While datasets from Italy exhibit selective high correlations between specific measurement pairs, such as PT and TS, the majority of sites demonstrate weak inter-site correlations. In contrast, Turkey reveals consistent positive correlations across all measurement site pairs. This uniform correlation pattern in the Turkish data may be attributed to the environmental conditions during data collection and geographical proximity, in addition to the the limited 24-hours sampling period, which could potentially inflate the observed correlation coefficients due to the shorter temporal window. 

\subsubsection{Cyclical Pattern Analysis}
We observe that all time series within the dataset from Italy exhibit cyclical patterns, as shown in Figure~\ref{fig:timeseries}. Using the FFT, we analyze these periodicities in the frequency domain, as illustrated in Figures~\ref{fig:dee22_fft} and \ref{fig:uh22_fft}, which display the magnitude of Fourier coefficients as a function of period (1/Hz) rather than frequency (Hz). This representation, where a red vertical line marks the 24-hours period, clearly reveals dominant peaks at both 12-hours and 24-hours duration, demonstrating strong daily periodic components in the EMF exposure signals across both DEE '22 and UH '22 environments. Similar observations were noted for other datasets from Italy.

\subsubsection{Stationarity of EMF Exposure}
\begin{figure}[ht]
    \centering
       \includegraphics[width=3.25in, totalheight=0.8\columnwidth]{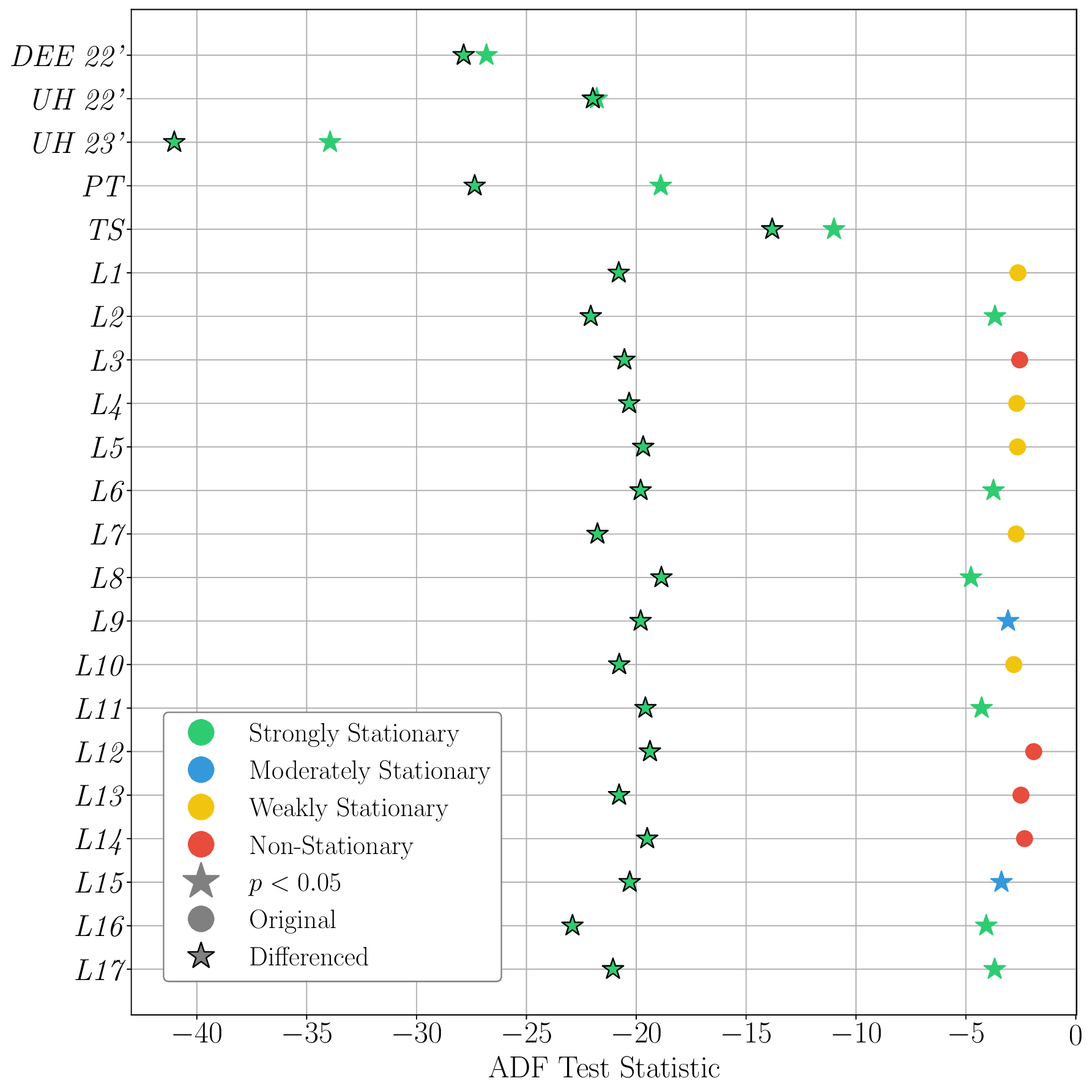}
    \caption{ADF Test to characterize stationarity of the  dataset from Turkey (17 locations) and Italy (4 locations). ADF test statistics and their statistical significance ($p$) are shown.} 
    \label{fig:adf}
\end{figure}

We apply the ADF test to assess stationarity, as shown in Figure~\ref{fig:adf}, which displays the ADF test statistics and statistical significance $(p)$ for the datasets considering two scenarios: the original raw data and first-order differenced data (indicated by dark borders). The results reveal that all time series from Italy are stationary ($p < 0.05$), while many measurement sites in Turkey exhibit non-stationary behavior. This distinction can be attributed to the limited 24-hours recording period in Turkey, which captures only a single daily cycle and thus appears as a non-stationary time series. In contrast, the Italy measurements, as demonstrated in the previous section, display cyclical patterns centered around a mean value without persistent trends, confirming their stationary nature. Upon applying first-order differencing, we observe improved (more negative) ADF statistics across both datasets, with all time series from Italy becoming increasingly stationary with significantly reduced ADF statistics.\footnote{ The stationarity of a time series, i.e., maintaining consistent statistical properties over time, suggests that windows drawn from it are more likely to be i.i.d., and thus exchangeable. This connection to exchangeability is particularly relevant for CP-based forecasting. This observation aligns with our empirical results, where we achieved state-of-the-art performance in both point forecasting and CP tasks on the stationary Italy datasets, while experiencing  modest results on the non-stationary Turkey dataset. 
}

\subsection{Training Configuration for DL Models}
We partition $(x_t)_{t = 1}^T$ into training, validation, and test datasets, with proportions of 70\%, 10\%, and 20\%, respectively.
For all considered DL models, including EMForecaster and baselines, we conduct training using the Adam optimizer across 100 epochs with a batch size of 2048, employing early stopping with a patience of 20 epochs on the validation Mean Square Error (MSE) \cite{adam}.
We use the MSE as our objective function and main evaluation metric on the test set, given by:
\begin{align} \label{MSE}
    \text{MSE}(\by, \hat{\by}) =  \frac{1}{O}\sum_{i=1}^{O} (y_{i} - \hat{y}_{i})^2,
\end{align}
for a prediction $\hat{\by} \in \R^{O}$ and label $\by \in \R^{O}$. We performed comprehensive hyperparameter tuning through grid search, optimizing separately for model, dataset, and prediction length $O$. Model training was executed on a single NVIDIA RTX 6000 Ada Generation GPU with 48GB of memory. The reported performance metrics on the test set represent averages across five random seeds, using the model configuration that achieved the lowest validation MSE (for the specific dataset and prediction length).




\begin{table*}[!htbp]
\centering
\caption{\small EMF point forecasting in terms of MSE with lookback window $L=336$ and prediction lengths $O \in \{96, 192, 336, 512\}$.}
\setlength{\tabcolsep}{6pt}
\begin{tabular}{llccccccccc}
\toprule
\small & \small  & \small EMForecaster & \small DLinear & \small TSMixer & \small PatchTST & \small Transformer & \small LSTM & \small CNN & \small MLP & \small ARIMA \\ 
\small & \small $O$ & \small (2025) & \small (2023) & \small (2023) & \small (2023) & \small (2017) & \small (1997) & \small (1988) & \small (1986) & \small (1970) \\
\midrule

\multirow{4}{*}{\rotatebox[origin=c]{90}{\small DEE 22'}} 
& \small $96$ & \small \textbf{0.3019} & \small 0.3488 & \small 0.3119 & \small 0.3710 & \small 0.7744 & \small 0.3409 & \small 0.3217 & \small 0.3217 & \small 1.2371 \\
& \small $192$ & \small \textbf{0.3461} & \small 0.4055 & \small 0.3561 & \small 0.4151 & \small 1.1312 & \small 0.4033 & \small 0.3794 & \small 0.3794 & \small 1.4484 \\
& \small $336$ & \small \textbf{0.3811} & \small 0.4791 & \small 0.3911 & \small 0.4664 & \small 1.1417 & \small 0.4426 & \small 0.4314 & \small 0.4314 & \small 1.4193 \\
& \small $512$ & \small \textbf{0.4436} & \small 0.5420 & \small 0.4536 & \small 0.5336 & \small 1.3673 & \small 0.4854 & \small 0.4806 & \small 0.4806 & \small 1.4333 \\

\midrule

\multirow{4}{*}{\rotatebox[origin=c]{90}{\small UH 22'}} 
& \small $96$ & \small \textbf{0.2404} & \small 0.2841 & \small 0.2685 & \small 0.2992 & \small 1.2874 & \small 0.3540 & \small 0.3599 & \small 0.4059 & \small 1.6604 \\
& \small $192$ & \small \textbf{0.2726} & \small 0.3159 & \small 0.3026 & \small 0.3345 & \small 1.2291 & \small 0.4863 & \small 0.4185 & \small 0.4175 & \small 1.9352 \\
& \small $336$ & \small \textbf{0.3148} & \small 0.3735 & \small 0.3580 & \small 0.3801 & \small 0.8394 & \small 0.5407 & \small 0.4733 & \small 0.4733 & \small 1.9328 \\
& \small $512$ & \small \textbf{0.3535} & \small 0.4165 & \small 0.4045 & \small 0.3973 & \small 0.6675 & \small 0.6049 & \small 0.5463 & \small 0.5463 & \small 1.9635 \\

\midrule

\multirow{4}{*}{\rotatebox[origin=c]{90}{\small UH 23'}} 
& \small $96$ & \small \textbf{0.2999} & \small 0.3251 & \small 0.3003 & \small 0.3490 & \small 1.0483 & \small 0.3180 & \small 0.3133 & \small 0.3133 & \small 1.0896 \\
& \small $192$ & \small \textbf{0.3158} & \small 0.3417 & \small \textbf{0.3158} & \small 0.4017 & \small 0.7665 & \small 0.3453 & \small 0.3391 & \small 0.3393 & \small 1.1964 \\
& \small $336$ & \small 0.3346 & \small 0.3716 & \small \textbf{0.3326} & \small 0.4083 & \small 0.6530 & \small 0.3517 & \small 0.3600 & \small 0.3600 & \small 1.2197 \\
& \small $512$ & \small 0.3434 & \small 0.3945 & \small \textbf{0.3429} & \small 0.4422 & \small 0.7776 & \small 0.3670 & \small 0.3787 & \small 0.3787 & \small 1.2328 \\

\midrule

\multirow{4}{*}{\rotatebox[origin=c]{90}{\small PT}} 
& \small $96$ & \small \textbf{0.1383} & \small 0.1423 & \small 0.1481 & \small 0.1460 & \small 0.3441 & \small 0.2027 & \small 0.1554 & \small 0.1554 & \small 0.2151 \\
& \small $192$ & \small \textbf{0.1476} & \small 0.1483 & \small 0.1564 & \small 0.1509 & \small 0.3970 & \small 0.2536 & \small 0.1616 & \small 0.1616 & \small 0.2931 \\
& \small $336$ & \small \textbf{0.1571} & \small 0.1592 & \small 0.1697 & \small 0.1632 & \small 0.3977 & \small 0.2546 & \small 0.1723 & \small 0.1723 & \small 0.3348 \\
& \small $512$ & \small \textbf{0.1656} & \small 0.1670 & \small 0.1756 & \small 0.1675 & \small 0.3314 & \small 0.2571 & \small 0.1787 & \small 0.1787 & \small 0.3574 \\

\midrule

\multirow{4}{*}{\rotatebox[origin=c]{90}{\small Turkey}} 
& \small $96$ & \small 0.4174 & \small \textbf{0.4054} & \small 0.4396 & \small 0.4822 & \small 0.5349 & \small 0.7714 & \small 0.7660 & \small 0.8014 & \small 0.8014 \\
& \small $192$ & \small 0.4564 & \small \textbf{0.4422} & \small 0.4793 & \small 0.5348 & \small 0.5804 & \small 0.8221 & \small 0.7758 & \small 0.8143 & \small 0.8143 \\
& \small $336$ & \small 0.4885 & \small \textbf{0.4818} & \small 0.5106 & \small 0.6050 & \small 0.6007 & \small 0.7917 & \small 0.7827 & \small 0.8296 & \small 0.8296 \\
& \small $512$ & \small 0.5230 & \small \textbf{0.5101}  & \small 0.5463 & \small 0.6612 & \small 0.6241 & \small 0.8018 & \small 0.7848 & \small 0.8356 & \small 0.8356 \\

\bottomrule
\end{tabular}
\label{table:point_forecasting}
\end{table*}



\section{Baselines, Results, and Discussions \label{sec:results}}
In this section, we describe the considered baselines and analyze the performance of the proposed EMForecaster for both \textit{point forecasting}, which provides deterministic predictions, and \textit{conformal forecasting}, which generates prediction intervals with statistical guarantees.

\subsection{Baselines}
The considered baselines are listed as follows: \\
$\bullet$ \textbf{MLP} processes the input time series through multiple fully-connected layers, capturing non-linear relationships between historical and future values through a direct mapping approach. \\
$\bullet$ \textbf{LSTM} uses gating mechanisms and recurrent connections to model temporal dependencies, enabling selective retention of relevant historical information across the series\cite{lstm}.\\
$\bullet$ \textbf{CNNs} utilize 1D convolutions sliding over the input sequence to extract local temporal patterns, effectively capturing hierarchical features through shared parameters.\\
$\bullet$ \textbf{Transformer} employs self-attention mechanisms to model relationships among all time steps simultaneously, allowing it to capture both long-range dependencies and local patterns. The parallel processing nature of attention makes it particularly effective at modeling complex temporal interactions \cite{transformer}. In this work, we use the standard encoder-only architecture. \\
$\bullet$ \textbf{DLinear} decomposes time series into trends and seasonal components, processing each through separate linear layers, providing an interpretable approach that can capture both long-term trends and seasonal patterns \cite{dlinear}. \\
$\bullet$ \textbf{TSMixer} is an MLP-based time series model, inspired by MLP-mixer \cite{mlpmixer, tsmixer}. TSMixer processes time series through parallel MLPs that separately handle temporal patterns (across time steps) and feature interactions (across measurements). \\
$\bullet$ \textbf{PatchTST} segments input time series into patches and processes them using a Transformer architecture, combining the benefits of local pattern extraction through patching with the global modeling capabilities of self-attention  \cite{patchtst}.\\
$\bullet$ \textbf{ARIMA} is a traditional statistical learning model, which combines differencing, autoregression, and moving average components to model linear relationships in stationary time series data \cite{arima}.

\subsection{Results and Discussions: Point Forecasting}
\subsubsection{Comparative Analysis of EMForecaster with Baselines} 
We evaluate the performance of point forecasting, i.e., forecasting individual future time points in the traditional sense on all datasets. Window sampling is performed on all 17 time series from Turkey, before taking the union of all windows to produce a full, multi-measurement site dataset. In contrast, we considered each dataset from Italy separately (DEE 22', UH 22', UH 23', PT) due to their distributional differences and weaker correlation compared to Turkey, as shown in Fig.~\ref{fig:turkey_heatmap}. 

For each dataset, we evaluate four prediction lengths $O \in \{96, 192, 336, 512\}$ independently, with a fixed lookback window of $L = 336$. The reported sampling rate is 6 min for Italy, thus the forecast ranges from 9.6 hrs to 51.2 hrs with a lookback window of 33.6 hrs. For the dataset from Turkey, the lookback window is fixed at 84 min, with prediction lengths ranging from 24 min to 128 min. Therefore, although the units for $L$ and $O$ are identical, the forecasting lengths differ as the reported sampling rate is different in the two datsets.


{Table~\ref{table:point_forecasting} displays point forecasting performance across all baselines. EMForecaster achieves consistently superior MSE performance compared to modern DL baselines (PatchTST, TSMixer, DLinear), while traditional approaches (Transformer, LSTM, CNN, MLP, and ARIMA) show higher MSE on average. While DLinear and TSMixer achieve comparable performance on the PT and Turkish datasets, their performance degrades significantly on other datasets, particularly with increasing forecast horizons $O$. EMForecaster demonstrates robust performance across the Italian datasets, though it shows slightly lower performance on the Turkish dataset due to inherent non-stationarity and limited data availability. These results indicate that in stationary environments, EMForecaster exhibits strong generalization capabilities across diverse settings and forecast horizons.}
\begin{figure*}[t]
    \centering
    \begin{subfigure}[b]{0.48\textwidth}
        \includegraphics[width=\textwidth]{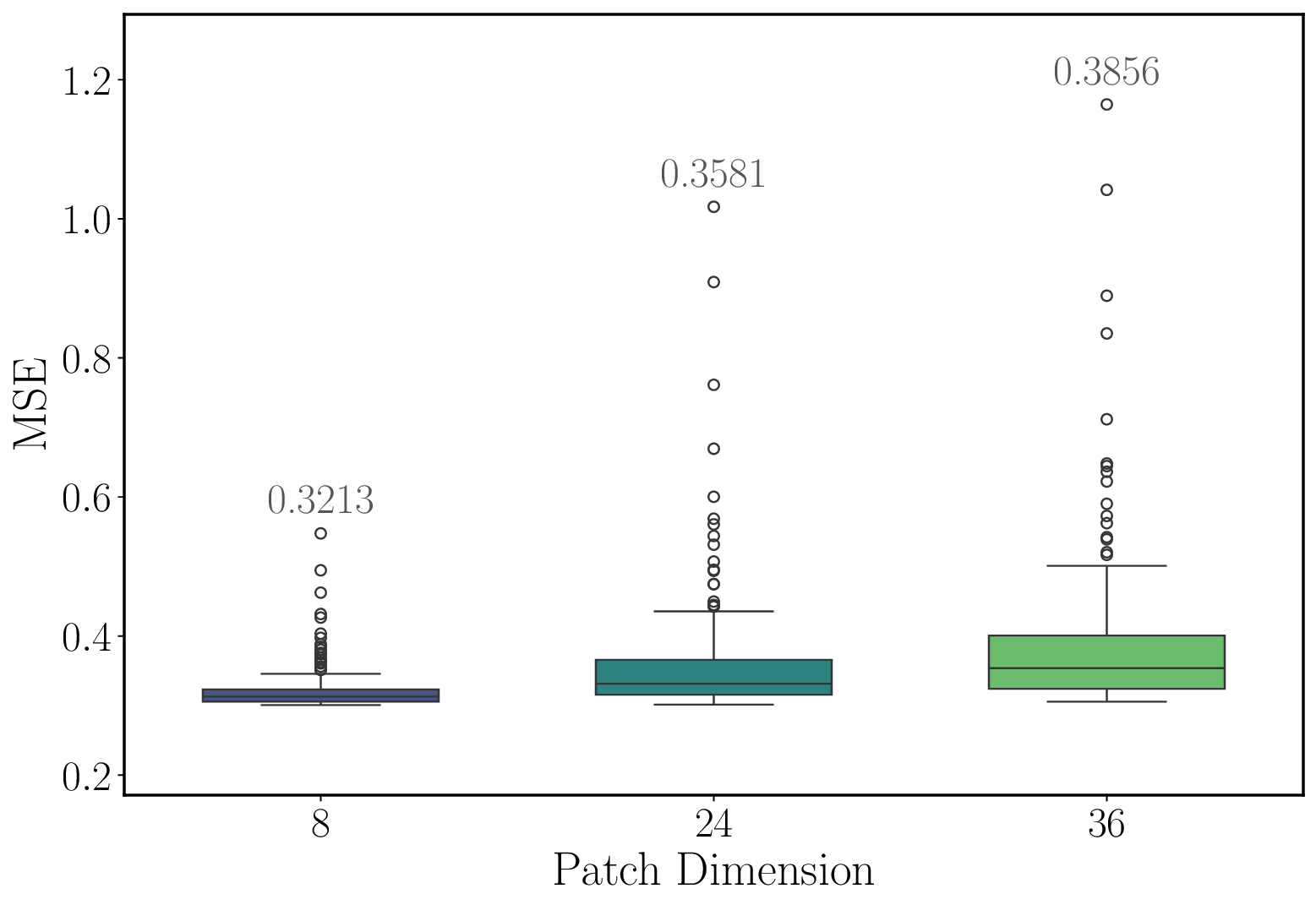}
    \end{subfigure}
    \hfill
    \begin{subfigure}[b]{0.48\textwidth}
        \includegraphics[width=\textwidth]{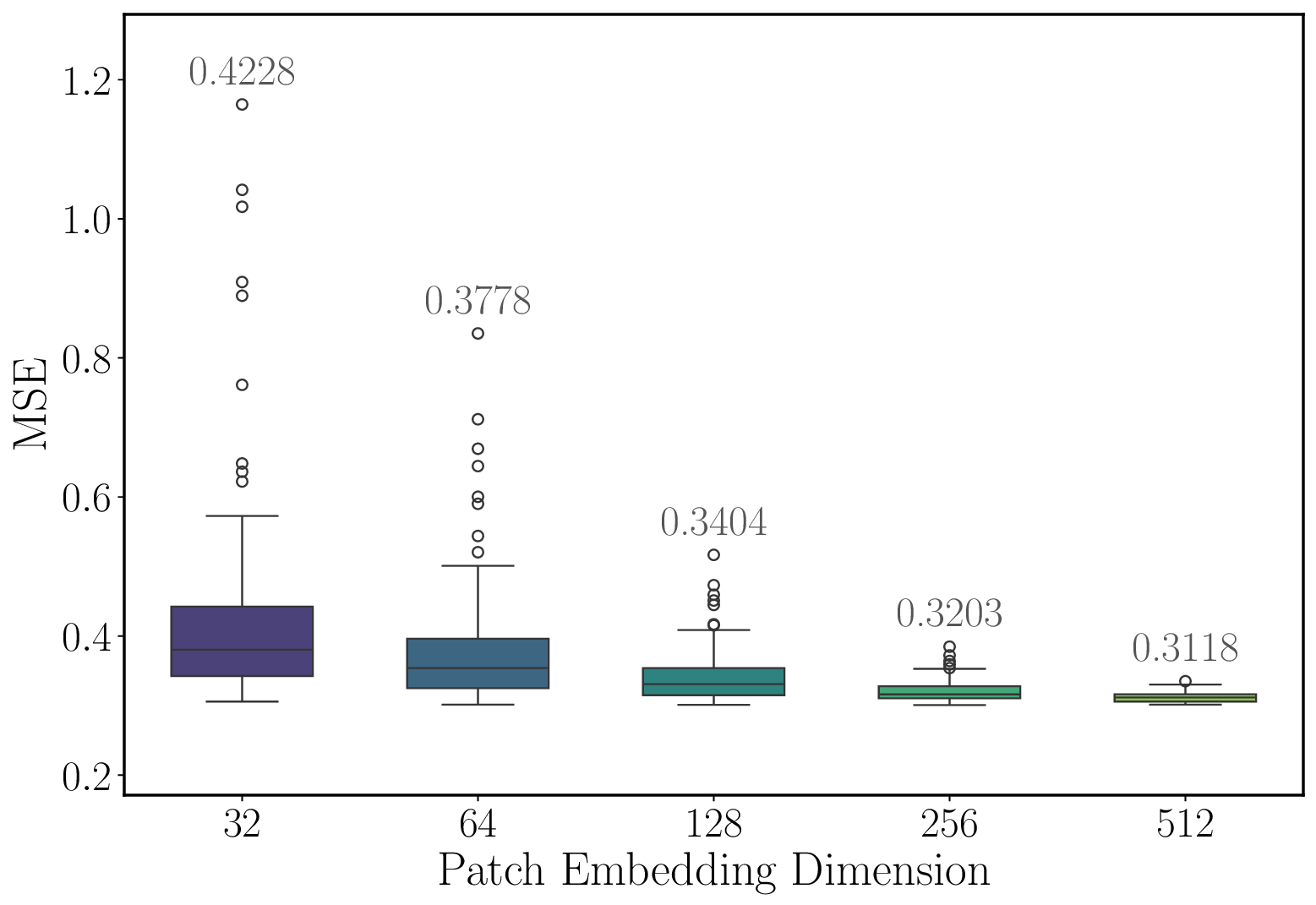}
    \end{subfigure}
    \caption{\small Boxplot of MSE for the patch dimension $P$ and patch embedding dimension $D$, over multiple different hyperparameter configurations; MSE reported at the top for each configuration.}
    \label{fig:ablations}
\end{figure*}

\subsubsection{Analysis of Patch Dimension and Patch Embedding Dimension}
In Figure~\ref{fig:ablations}, we analyze how the patch dimension ($P$) and patch embedding dimension ($D$) affect EMForecaster's performance through MSE scores, averaged across hyperparameter configurations and random seeds. Our results reveal distinct patterns: increasing $P$ leads to higher MSE (worse performance), while increasing $D$ consistently reduces MSE (better performance).
The inverse relationship between patch size and performance suggests that smaller patches enable more effective representation learning by providing focused, localized views of the time series. Larger patches may introduce redundant information that obscures relevant temporal patterns. Conversely, larger embedding dimensions ($D$) enhance model performance by providing greater representational capacity for each patch. This increased dimensionality allows the patch MLP and time MLP components to learn more expressive mappings, enabling better adaptation to diverse temporal patterns. The empirical benefits of $D \gg P$ support our architectural choice of expanding patch representations through the embedding layer.

\subsection{Results and Discussions: Conformal Forecasting}
We conduct our conformal forecasting experiments on the same datasets used for point forecasting for the prediction length $O = 96$ with lookback window $L = 336$ using the framework in section~\ref{sec:ctsf} on EMForecaster and the baselines. 

\begin{figure}[t]
    \centering
    \includegraphics[width=\columnwidth]{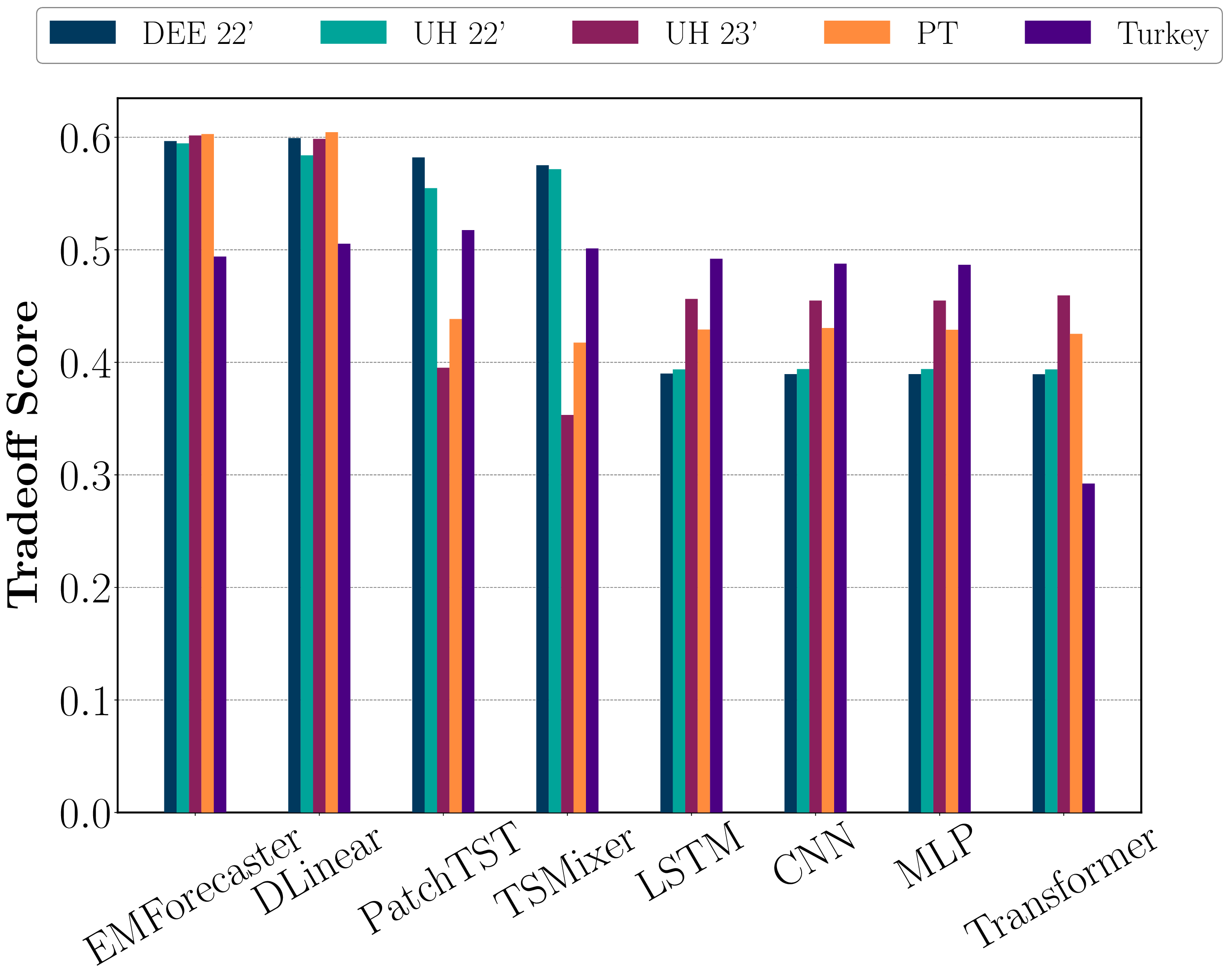}
    \caption{Performance comparison of different forecasting models across datasets using the Tradeoff Score (TOS) metric using $\alpha=0.01$, $\beta = \frac{2}{3}$, and $\lambda = \frac{1}{2}$.}
    \label{fig:tos}
\end{figure}

\begin{figure}[t]
    \centering
        \includegraphics[width=0.8\columnwidth]{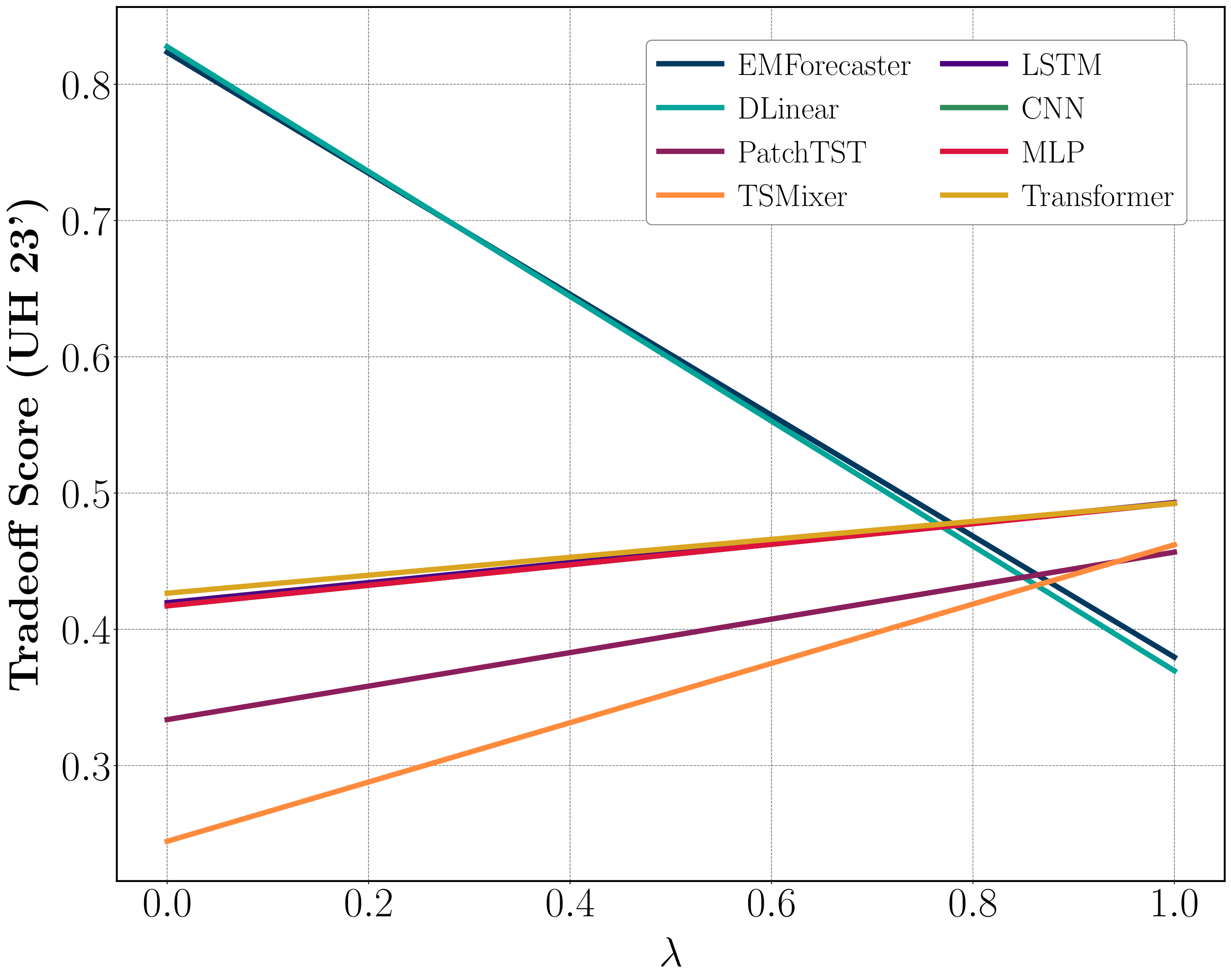}
    \caption{Tradeoff Score (TOS) as a function of $\lambda$ with 
$\alpha = 0.01$ and $\beta = \frac{2}{3}$ for UH 23' dataset.}
    \label{fig:lambda_analysis_combined}
\end{figure}

\begin{figure}[t]
    \centering
        \includegraphics[width=0.8\columnwidth]{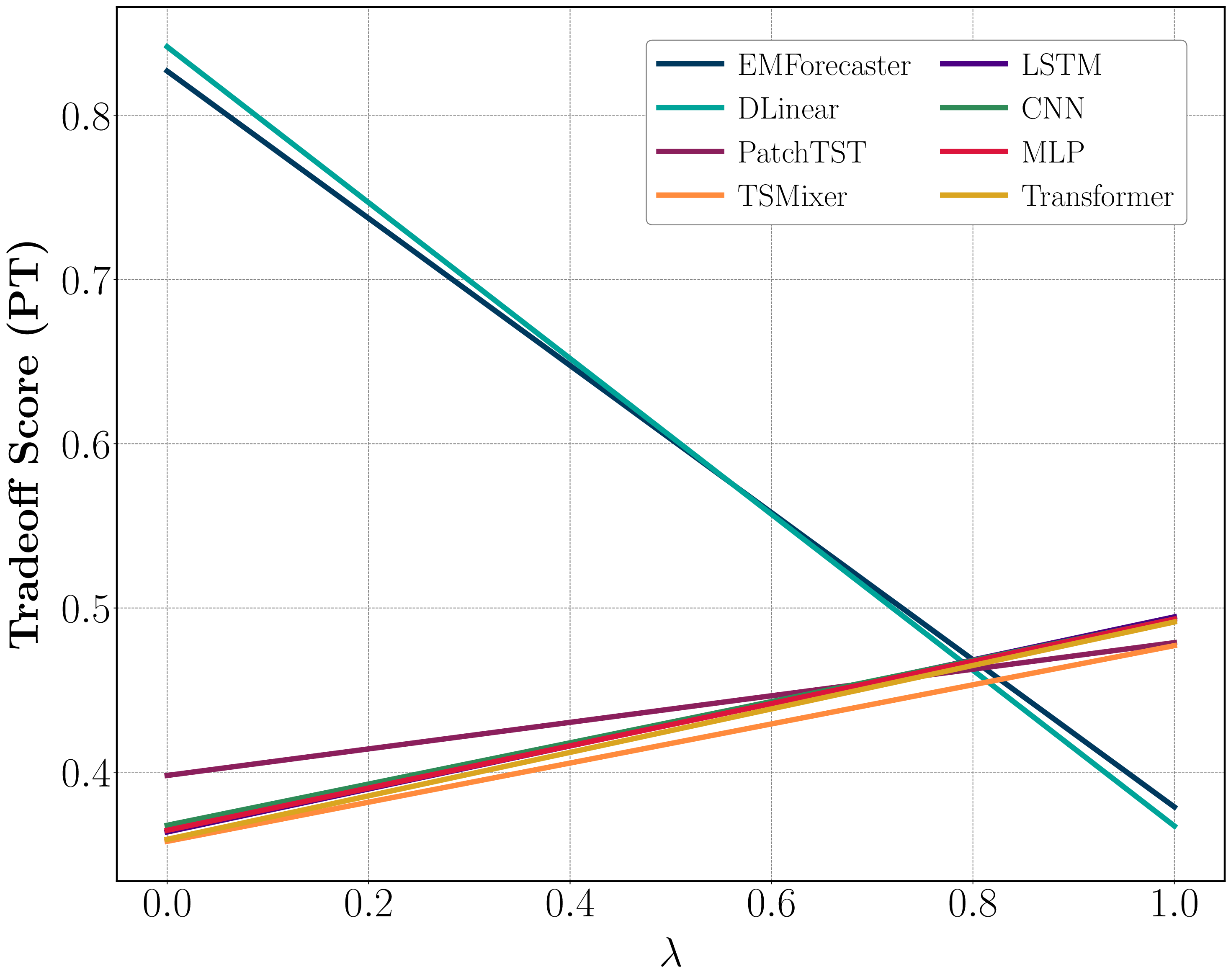}
    \caption{Tradeoff Score (TOS) as a function of $\lambda$ with 
$\alpha = 0.01$ and $\beta = \frac{2}{3}$ for PT dataset.}
    \label{fig:lambda_analysis_combined2}
\end{figure}

\subsubsection{Comparative Analysis of EMForecaster with CP}
Fig.~\ref{fig:tos} displays the results across all datasets for each model with respect to the TOS. We see that EMForecaster achieves comparable performance to DLinear while achieving superior performance compared to all other baselines. For this experiment, we used $\beta = \frac{2}{3}$ and $\lambda = \frac{1}{2}$, as suggested in the discussion from section~\ref{sec:tos}. $\beta = \frac{2}{3}$ skews the WAC to the joint coverage as it is a harder metric to optimize, while still optimizing partly for the IC, while $\lambda = \frac{1}{2}$ provides an equal importance between coverage and prediction interval width when computing the TOS. 
 
\subsubsection{Choice of $\lambda$ in Conformal Forecasting}
Fig.~\ref{fig:lambda_analysis_combined} and Fig.~\ref{fig:lambda_analysis_combined2} examine the impact of $\lambda$ on the TOS  for the UH 23' and PT datasets respectively. Both EMForecaster and DLinear outperform all other baselines up to $\lambda \approx 0.8$. Beyond this point, the TOS becomes heavily influenced by coverage. 
Thus, while many baselines can achieve good coverage, it requires a very large $\lambda$ to achieve an optimal TOS, indicating a poor balance between obtaining coverage and interval widths for most baselines. In contrast, EMForecaster and DLinear strike an optimal balance, avoiding excessively wide intervals while maintaining strong coverage.

\subsection{Analysis of Differencing and Temporal Resolution}

Table~\ref{table:differencing_and_temporal} presents  EMForecaster's performance across different temporal resolutions. We compare results at the original sampling rate of $\Delta t$ = 6 min and a down-sampled rate of $\Delta t$ = 30 min, where the latter is obtained by averaging five consecutive 6 min measurements. While independent coverage remains consistent across both sampling rates, we observe a notable decrease in joint coverage for the UH '22 dataset at $\Delta t$ = 30 min. The coarser temporal resolution ($\Delta t$ = 30 min) generally yields lower MSE due to the smoothing effect that reduces high-frequency noise. Interestingly, despite higher MSE values, the $\Delta t$ = 6 min resolution achieves narrower mean prediction intervals, suggesting superior conformal forecasting performance. This demonstrates a trade-off between point forecast accuracy and the precision of uncertainty quantification across different temporal resolutions.

\begin{table}[!htbp]
\centering
\caption{\small Performance evaluation of EMForecaster, examining the effect of sampling rates $\Delta t$ across several datasets, with lookback window $L=336$ and prediction length $O = 96$.}

\setlength{\tabcolsep}{3pt}
\begin{subtable}{\linewidth}
\centering
\begin{tabular}{@{}l@{\hspace{2pt}}l@{\hspace{2pt}}c@{\hspace{4pt}}c@{\hspace{4pt}}c@{\hspace{4pt}}c@{}}
\toprule
\small & \small \ \  \ $\Delta t$ & \small MSE & \small JC & \small IC & \small MIW \\
\midrule
\vspace{0.1cm}
\multirow{2}{*}{\rotatebox[origin=c]{90}{\scriptsize  DEE 22'}} 
& \small \ \ 6min  \ & \small 0.3008  & \small 82.89  & \small 98.98  & \small 4.99  \\
& \small \ \  30min & \small 0.2104 & \small 84.96 & \small 99.49 & \small 6.93 \\
\midrule
\vspace{0.1cm}
\multirow{2}{*}{\rotatebox[origin=c]{90}{\scriptsize UH 22'}} 
& \small \ \  6min \ & \small 0.2467 & \small 73.70 & \small 97.95 & \small 4.11 \\
& \small \ \  30min \ & \small 0.2133 & \small 51.19 & \small 96.07 & \small 5.94 \\
\midrule
\vspace{0.1cm}
\multirow{2}{*}{\rotatebox[origin=c]{90}{\scriptsize PT}} 
& \small \ \ 6min  \ & \small 0.1372 & \small 62.41 & \small 98.58 & \small 2.73 \\
& \small \ \  30min \ & \small 0.1181 & \small 65.04 & \small 98.81 & \small 3.09 \\
\bottomrule
\end{tabular}
\end{subtable}


\label{table:differencing_and_temporal}
\end{table}
\subsection{Computational Complexity}
For DET 22' with lookback window $L = 336$, forecast horizon $O = 96$, and $\alpha = 0.01$, the combined point forecasting and conformal forecasting training and evaluation durations were 38.51 sec for EMForecaster, 27.19 sec for DLinear, 10.98 min for TSMixer, 10.87 min for PatchTST, 1.25 min for the Transformer, 1.04 min for the LSTM, 52.68 sec for the MLP, and 54.99 sec for the CNN. Our computational results demonstrate that EMForecaster is the second fastest model, with only DLinear achieving marginally faster execution times. Modern architectures such as TSMixer and PatchTST exhibit significantly longer training times, approximately 17 times that of EMForecaster, indicating less efficient parameter utilization. In addition, EMForecaster outperforms traditional DL architectures including LSTM and Transformer in terms of computational efficiency, executing 1.6-2x faster, which makes it particularly suitable for applications requiring rapid training or deployment under computational constraints.



\section{Conclusion \label{sec:conclusion}}

In this work, we have introduced EMForecaster, a DL-empowered time series forecasting framework designed to predict EMF trends across diverse locations and varying forecast horizons beyond 50 hours. To improve the reliability of EMF predictions, we integrated a distribution-free uncertainty quantification framework, using CP, ensuring the ground truth falls within a specified prediction interval with a user-specified error rate.  To enhance the evaluation of conformal forecasting, we introduce a novel TOS evaluation metric which balances prediction interval width and empirical coverage, allowing for objective comparisons across different conformal predictors. Our extensive experiments demonstrate that EMForecaster significantly outperforms existing DL models in both point forecasting and conformal forecasting. Future research directions include extending EMForecaster to Beyond 5G (B5G) data and expanding the forecast horizon to multiple years.


\section{Acknowledgments}
We sincerely appreciate Dr. Sara Adda from Agenzia Regionale Protezione Ambiente Piemonte (ARPA Piemonte), Ivrea, Italy, and Dr. Cetin Kurnaz from the Department of Electrical and Electronics Engineering, Ondokuz Mayis University, Turkey, for generously providing valuable datasets that contributed to this study. While the majority of our results are based on datasets from the Department of Electronic Engineering at the University of Rome Tor Vergata and University Hospital in Rome, the datasets from Dr. Adda published in \cite{adda2023highnoon} and Dr. Kurnaz  published in \cite{kurnaz2020rfemf} were used for benchmarking purposes. 
Interested researchers are encouraged to contact the respective authors for data access.


\bibliography{bibtex}

\begin{thebibliography}{10}
\providecommand{\url}[1]{#1}
\csname url@samestyle\endcsname
\providecommand{\newblock}{\relax}
\providecommand{\bibinfo}[2]{#2}
\providecommand{\BIBentrySTDinterwordspacing}{\spaceskip=0pt\relax}
\providecommand{\BIBentryALTinterwordstretchfactor}{4}
\providecommand{\BIBentryALTinterwordspacing}{\spaceskip=\fontdimen2\font plus
\BIBentryALTinterwordstretchfactor\fontdimen3\font minus \fontdimen4\font\relax}
\providecommand{\BIBforeignlanguage}[2]{{%
\expandafter\ifx\csname l@#1\endcsname\relax
\typeout{** WARNING: IEEEtran.bst: No hyphenation pattern has been}%
\typeout{** loaded for the language `#1'. Using the pattern for}%
\typeout{** the default language instead.}%
\else
\language=\csname l@#1\endcsname
\fi
#2}}
\providecommand{\BIBdecl}{\relax}
\BIBdecl

\bibitem{amin}
M.~A. Saeidi, H.~Tabassum, and M.~Alizadeh, ``Molecular absorption-aware user assignment, spectrum, and power allocation in dense {THz} networks with multi-connectivity,'' \emph{IEEE Trans. on Wireless Commun.}, pp. 1--1, 2024.

\bibitem{9518367}
L.~Chiaraviglio, A.~Elzanaty, and M.-S. Alouini, ``Health risks associated with 5{G} exposure: A view from the communications engineering perspective,'' \emph{IEEE Open Jrnl. of the Commun. Society}, vol.~2, pp. 2131--2179, 2021.

\bibitem{ICNIRP}
\BIBentryALTinterwordspacing
(ICNIRP), ``Guidelines for limiting exposure to electromagnetic fields (100 {kH}z to 300 {GH}z),'' \emph{Health Physics}, 2020. [Online]. Available: \url{https://www.icnirp.org/cms/upload/publications/ICNIRPrfgdl2020.pdf}
\BIBentrySTDinterwordspacing

\bibitem{stam2018comparison}
R.~Stam, ``Comparison of international policies on electromagnetic fields:(power frequency and radiofrequency fields),'' 2018.

\bibitem{app10238753}
\BIBentryALTinterwordspacing
M.~Al~Hajj, S.~Wang, L.~Thanh~Tu, S.~Azzi, and J.~Wiart, ``A statistical estimation of {{5G}} massive {MIMO} networks’ exposure using stochastic geometry in mmwave bands,'' \emph{Applied Sciences}, vol.~10, no.~23, 2020. [Online]. Available: \url{https://www.mdpi.com/2076-3417/10/23/8753}
\BIBentrySTDinterwordspacing

\bibitem{gontier2024impactdynamicbeamformingemf}
\BIBentryALTinterwordspacing
Q.~Gontier, C.~Wiame, J.~Wiart, F.~Horlin, C.~Tsigros, C.~Oestges, and P.~D. Doncker, ``On the impact of dynamic beamforming on {EMF} exposure and network coverage: A stochastic geometry perspective,'' 2024. [Online]. Available: \url{https://arxiv.org/abs/2405.01190}
\BIBentrySTDinterwordspacing

\bibitem{9511258}
N.~A. Muhammad, N.~Seman, N.~I.~A. Apandi, C.~T. Han, Y.~Li, and O.~Elijah, ``Stochastic geometry analysis of electromagnetic field exposure in coexisting sub-6 {GHz} and millimeter wave networks,'' \emph{IEEE Access}, vol.~9, pp. 112\,780--112\,791, 2021.

\bibitem{gontier2024uplinkdownlinkemfexposure}
\BIBentryALTinterwordspacing
Q.~Gontier, C.~Wiame, J.~Wiart, F.~Horlin, C.~Tsigros, C.~Oestges, and P.~D. Doncker, ``On the uplink and downlink {EMF} exposure and coverage in dense cellular networks: A stochastic geometry approach,'' 2024. [Online]. Available: \url{https://arxiv.org/abs/2312.08978}
\BIBentrySTDinterwordspacing

\bibitem{10536047}
Q.~Gontier, C.~Wiame, S.~Wang, M.~Di~Renzo, J.~Wiart, F.~Horlin, C.~Tsigros, C.~Oestges, and P.~De~Doncker, ``Joint metrics for {EMF} exposure and coverage in real-world homogeneous and inhomogeneous cellular networks,'' \emph{IEEE Trans. on Wireless Commun}, vol.~23, no.~10, pp. 13\,267--13\,284, 2024.

\bibitem{10504892}
Y.~Qin, M.~A. Kishk, A.~Elzanaty, L.~Chiaraviglio, and M.-S. Alouini, ``Unveiling passive and active {EMF} exposure in large-scale cellular networks,'' \emph{IEEE Open Jrnl. of the Commun Society}, vol.~5, pp. 2991--3006, 2024.

\bibitem{10047969}
L.~Chen, A.~Elzanaty, M.~A. Kishk, L.~Chiaraviglio, and M.-S. Alouini, ``Joint uplink and downlink {EMF} exposure: Performance analysis and design insights,'' \emph{IEEE Trans. on Wireless Commun}, vol.~22, no.~10, pp. 6474--6488, 2023.

\bibitem{9462948}
Q.~Gontier, L.~Petrillo, F.~Rottenberg, F.~Horlin, J.~Wiart, C.~Oestges, and P.~De~Doncker, ``A stochastic geometry approach to {EMF} exposure modeling,'' \emph{IEEE Access}, vol.~9, pp. 91\,777--91\,787, 2021.

\bibitem{10225716}
C.~Wiame, S.~Demey, L.~Vandendorpe, P.~De~Doncker, and C.~Oestges, ``Joint data rate and {EMF} exposure analysis in manhattan environments: Stochastic geometry and ray tracing approaches,'' \emph{IEEE Trans. on Veh. Tech.}, vol.~73, no.~1, pp. 894--908, 2024.

\bibitem{sambo2016electromagnetic}
Y.~A. Sambo, M.~Al-Imari, F.~H{\'e}liot, and M.~A. Imran, ``Electromagnetic emission-aware schedulers for the uplink of {OFDM} wireless communication systems,'' \emph{IEEE Trans. on Veh. Tech.}, vol.~66, no.~2, pp. 1313--1323, 2016.

\bibitem{jiang2023rate}
H.~Jiang, L.~You, A.~Elzanaty, J.~Wang, W.~Wang, X.~Gao, and M.-S. Alouini, ``Rate-splitting multiple access for uplink massive {MIMO} with electromagnetic exposure constraints,'' \emph{IEEE Jrnl. on Sel. areas in Commun.}, vol.~41, no.~5, pp. 1383--1397, 2023.

\bibitem{sambo2014user}
Y.~A. Sambo, F.~H{\'e}liot, and M.~A. Imran, ``A user scheduling scheme for reducing electromagnetic emission in the uplink of mobile communication systems,'' in \emph{IEEE Online Conf. on Green Commun. (OnlineGreenComm)}, 2014, pp. 1--5.

\bibitem{sambo2017electromagnetic}
Y.~A. Sambo, F.~Heliot, and M.~A. Imran, ``Electromagnetic emission-aware scheduling for the uplink of multicell {OFDM} wireless systems,'' \emph{IEEE Trans. on Veh. Tech.}, vol.~66, no.~9, pp. 8212--8222, 2017.

\bibitem{matalatala2018joint}
M.~Matalatala, M.~Deruyck, E.~Tanghe, S.~Goudos, L.~Martens, and W.~Joseph, ``Joint optimization towards power consumption and electromagnetic exposure for massive {MIMO} {5G} networks,'' in \emph{IEEE 29th Annual Intl. Symposium on Personal, Indoor and Mobile Radio Commun. (PIMRC)}, 2018, pp. 1208--1214.

\bibitem{wang2011evaluation}
M.~Wang, L.~Lin, J.~Chen, D.~Jackson, W.~Kainz, Y.~Qi, and P.~Jarmuszewski, ``Evaluation and optimization of the specific absorption rate for multi-antenna systems,'' \emph{IEEE Trans. on electromagnetic compatibility}, vol.~53, no.~3, pp. 628--637, 2011.

\bibitem{ying2013beamformer}
D.~Ying, D.~J. Love, and B.~M. Hochwald, ``Beamformer optimization with a constraint on user electromagnetic radiation exposure,'' in \emph{47th Annual Conf. on Information Sciences and Sys.(CISS)}, 2013, pp. 1--6.

\bibitem{penhoat2015enabling}
J.~Penhoat, R.~Ag{\"u}ero, F.~Heliot, and M.~Tesanovic, ``Enabling low electromagnetic exposure multimedia sessions on an {LTE} network with an {IP} multimedia subsystem control plane,'' in \emph{Mobile Networks and Management: 6th Intl. Conf., MONAMI}.\hskip 1em plus 0.5em minus 0.4em\relax Springer, 2015, pp. 207--216.

\bibitem{diez2015reducing}
L.~Diez, R.~Aguero, and J.~Penhoat, ``Reducing the electromagnetic exposure over {LTE} networks by means of an adaptive retransmission scheme: A use case based on a video service,'' in \emph{IEEE 81st Veh. Tech. Conf. (VTC Spring)}, 2015, pp. 1--5.

\bibitem{chiaraviglio2018planning}
L.~Chiaraviglio, A.~S. Cacciapuoti, G.~Di~Martino, M.~Fiore, M.~Montesano, D.~Trucchi, and N.~B. Melazzi, ``Planning {5G} networks under {EMF} constraints: State of the art and vision,'' \emph{IEEE Access}, vol.~6, pp. 51\,021--51\,037, 2018.

\bibitem{ITU:2019}
{International Telecommunication Union}, ``Guidance on complying with limits for human exposure to electromagnetic fields,'' \url{https://www.itu.int/rec/T-REC-K.Sup14-201909-I}, 2019, accessed: 2025-01-02.

\bibitem{matalatala2019multi}
M.~Matalatala, M.~Deruyck, S.~Shikhantsov, E.~Tanghe, D.~Plets, S.~Goudos, K.~E. Psannis, L.~Martens, and W.~Joseph, ``Multi-objective optimization of massive {MIMO} {5G} wireless networks towards power consumption, uplink and downlink exposure,'' \emph{Applied Sciences}, vol.~9, no.~22, p. 4974, 2019.

\bibitem{matalatala2018optimal}
M.~Matalatala, M.~Deruyck, E.~Tanghe, L.~Martens, and W.~Joseph, ``Optimal low-power design of a multicell multiuser massive {MIMO} system at 3.7 {GHz} for {5G} wireless networks,'' \emph{Wireless Commun. and Mobile Computing}, vol. 2018, no.~1, p. 9796784, 2018.

\bibitem{amaldi2003planning}
E.~Amaldi, A.~Capone, and F.~Malucelli, ``Planning {UMTS} base station location: Optimization models with power control and algorithms,'' \emph{IEEE Trans. on Wireless Commun.}, vol.~2, no.~5, pp. 939--952, 2003.

\bibitem{oughton2019open}
E.~J. Oughton, K.~Katsaros, F.~Entezami, D.~Kaleshi, and J.~Crowcroft, ``An open-source techno-economic assessment framework for {5G} deployment,'' \emph{IEEE Access}, vol.~7, pp. 155\,930--155\,940, 2019.

\bibitem{miller2024survey}
\BIBentryALTinterwordspacing
J.~A. Miller, M.~Aldosari, F.~Saeed, N.~H. Barna, S.~Rana, I.~B. Arpinar, and N.~Liu, ``A survey of deep learning and foundation models for time series forecasting,'' 2024. [Online]. Available: \url{https://arxiv.org/abs/2401.13912}
\BIBentrySTDinterwordspacing

\bibitem{arima}
G.~E. Box, G.~M. Jenkins, G.~C. Reinsel, and G.~M. Ljung, \emph{Time series analysis: forecasting and control}.\hskip 1em plus 0.5em minus 0.4em\relax John Wiley \& Sons, 2015.

\bibitem{kiouvrekis2024comparative}
Y.~Kiouvrekis, I.~Givisis, T.~Panagiotakopoulos, I.~Tsilikas, A.~Ploussi, E.~Spyratou, and E.~P. Efstathopoulos, ``A comparative analysis of explainable artificial intelligence models for electric field strength prediction over eight european cities,'' \emph{Sensors}, vol.~25, no.~1, p.~53, 2024.

\bibitem{bakcan2022measurement}
M.~R. Bakcan \emph{et~al.}, ``Measurement and prediction of electromagnetic radiation exposure level in a university campus,'' \emph{Tehnivcki vjesnik}, vol.~29, no.~2, pp. 449--455, 2022.

\bibitem{pala2021examining}
Z.~Pala, ``Examining {EMF} time series using prediction algorithms with {R},'' \emph{IEEE Canadian Jrnl. of Electrical and Computer Engineering}, vol.~44, no.~2, pp. 223--227, 2021.

\bibitem{nguyen2024deep}
C.~Nguyen, A.~A. Cheema, C.~Kurnaz, A.~Rahimian, C.~Brennan, and T.~Q. Duong, ``Deep learning models for time-series forecasting of {RF-EMF} in wireless networks,'' \emph{IEEE Open Jrnl. of the Commun. Soc.}, 2024.

\bibitem{transformer}
A.~Vaswani, N.~Shazeer, N.~Parmar, J.~Uszkoreit, L.~Jones, A.~N. Gomez, {\L}.~Kaiser, and I.~Polosukhin, ``Attention is all you need,'' \emph{Advances in neural information processing systems}, vol.~30, 2017.

\bibitem{kurnaz2020rfemf}
C.~Kurnaz and M.~Mutlu, ``Comprehensive radiofrequency electromagnetic field measurements and assessments: a city center example,'' \emph{Environmental Monitoring and Assessment}, vol. 192, pp. 1--14, 2020.

\bibitem{liu2022non}
Y.~Liu, H.~Wu, J.~Wang, and M.~Long, ``Non-stationary transformers: Exploring the stationarity in time series forecasting,'' \emph{Advances in Neural Information Processing Systems}, vol.~35, pp. 9881--9893, 2022.

\bibitem{santoro2024comparison}
D.~Santoro, T.~Ciano, and M.~Ferrara, ``A comparison between machine and deep learning models on high stationarity data,'' \emph{Scientific Reports}, vol.~14, no.~1, p. 19409, 2024.

\bibitem{adf}
S.~E. Said and D.~A. Dickey, ``Testing for unit roots in autoregressive-moving average models of unknown order,'' \emph{Biometrika}, vol.~71, no.~3, pp. 599--607, 1984.

\bibitem{enbpi}
C.~Xu and Y.~Xie, ``Conformal prediction interval for dynamic time-series,'' in \emph{Intl. Conf. on Machine Learning}.\hskip 1em plus 0.5em minus 0.4em\relax PMLR, 2021, pp. 11\,559--11\,569.

\bibitem{ctsf}
K.~Stankeviciute, A.~M~Alaa, and M.~van~der Schaar, ``Conformal time-series forecasting,'' \emph{Advances in neural information processing systems}, vol.~34, pp. 6216--6228, 2021.

\bibitem{bloomfield2004fourier}
P.~Bloomfield, \emph{Fourier analysis of time series: an introduction}.\hskip 1em plus 0.5em minus 0.4em\relax John Wiley \& Sons, 2004.

\bibitem{nikravesh2016mobile}
A.~Y. Nikravesh, S.~A. Ajila, C.-H. Lung, and W.~Ding, ``{Mobile network traffic prediction using MLP, MLPWD, and SVM},'' in \emph{IEEE Intl. congress on big data (BigData Congress)}, 2016, pp. 402--409.

\bibitem{di2023multivariate}
M.~Di~Mauro, G.~Galatro, F.~Postiglione, W.~Song, and A.~Liotta, ``Multivariate time series characterization and forecasting of {VoIP} traffic in real mobile networks,'' \emph{IEEE Trans. on Network and Service Management}, vol.~21, no.~1, pp. 851--865, 2023.

\bibitem{dalgkitsis2018traffic}
A.~Dalgkitsis, M.~Louta, and G.~T. Karetsos, ``Traffic forecasting in cellular networks using the {LSTM RNN},'' in \emph{Proc.of the 22nd Pan-Hellenic Conf. on informatics}, 2018, pp. 28--33.

\bibitem{habib2024transformer}
M.~A. Habib, P.~E.~I. Rivera, Y.~Ozcan, M.~Elsayed, M.~Bavand, R.~Gaigalas, and M.~Erol-Kantarci, ``Transformer-based wireless traffic prediction and network optimization in {O-RAN},'' in \emph{2024 IEEE Intl. Conf. on Commun. Workshops (ICC Workshops)}.\hskip 1em plus 0.5em minus 0.4em\relax IEEE, 2024, pp. 1--6.

\bibitem{perez2024dissecting}
P.~F. Perez, C.~Fiandrino, M.~Fiore, and J.~Widmer, ``Dissecting advanced time series forecasting models with aichronolens,'' in \emph{IEEE INFOCOM 2024-IEEE Conf. on Computer Commun. Workshops (INFOCOM WKSHPS)}.\hskip 1em plus 0.5em minus 0.4em\relax IEEE, 2024, pp. 01--02.

\bibitem{isravel2024multivariate}
D.~P. Isravel, S.~Silas, J.~W. Kathrine, E.~B. Rajsingh, and J.~Andrew, ``Multivariate forecasting of network traffic in {SDN} based ubiquitous healthcare system,'' \emph{IEEE Open Journal of the Communications Society}, 2024.

\bibitem{sone2020wireless}
S.~P. Sone, J.~J. Lehtom{\"a}ki, and Z.~Khan, ``Wireless traffic usage forecasting using real enterprise network data: Analysis and methods,'' \emph{IEEE Open Journal of the Communications Society}, vol.~1, pp. 777--797, 2020.

\bibitem{zhang2021deep}
Y.~Zhang, Y.~Wu, A.~Liu, X.~Xia, T.~Pan, and X.~Liu, ``Deep learning-based channel prediction for leo satellite massive mimo communication system,'' \emph{IEEE Wireless Commun. Letters}, vol.~10, no.~8, pp. 1835--1839, 2021.

\bibitem{hameed2022toward}
A.~Hameed, J.~Violos, A.~Leivadeas, N.~Santi, R.~Gr{\"u}nblatt, and N.~Mitton, ``{Toward QoS prediction based on temporal transformers for IoT applications},'' \emph{IEEE Trans. on Network and Service Management}, vol.~19, no.~4, pp. 4010--4027, 2022.

\bibitem{colpitts2023short}
A.~G. Colpitts and B.~R. Petersen, ``Short-term multivariate {KPI} forecasting in rural fixed wireless {LTE} networks,'' \emph{IEEE Networking Letters}, vol.~5, no.~1, pp. 11--15, 2023.

\bibitem{ak2021forecasting}
E.~Ak and B.~Canberk, ``Forecasting quality of service for next-generation data-driven wifi6 campus networks,'' \emph{IEEE Transactions on Network and Service Management}, vol.~18, no.~4, pp. 4744--4755, 2021.

\bibitem{zhou2022deep}
T.~Zhou, H.~Zhang, B.~Ai, C.~Xue, and L.~Liu, ``Deep-learning-based spatial--temporal channel prediction for smart high-speed railway communication networks,'' \emph{IEEE Transactions on Wireless Communications}, vol.~21, no.~7, pp. 5333--5345, 2022.

\bibitem{patchtst}
Y.~Nie, N.~H.~Nguyen, P.~Sinthong, and J.~Kalagnanam, ``A time series is worth 64 words: Long-term forecasting with transformers,'' in \emph{Intl. Conf. on Learning Representations}, 2023.

\bibitem{dlinear}
A.~Zeng, M.~Chen, L.~Zhang, and Q.~Xu, ``Are transformers effective for time series forecasting?'' in \emph{Proc. of the AAAI Conf. on Artificial Intelligence}, 2023.

\bibitem{cnn1}
K.~Fukushima, ``Neocognitron: A hierarchical neural network capable of visual pattern recognition,'' \emph{Neural networks}, vol.~1, no.~2, pp. 119--130, 1988.

\bibitem{cnn2}
Y.~LeCun, L.~Bottou, Y.~Bengio, and P.~Haffner, ``Gradient-based learning applied to document recognition,'' \emph{Proc. of the IEEE}, vol.~86, no.~11, pp. 2278--2324, 1998.

\bibitem{bai2018empirical}
S.~Bai, J.~Z. Kolter, and V.~Koltun, ``An empirical evaluation of generic convolutional and recurrent networks for sequence modeling,'' \emph{arXiv preprint arXiv:1803.01271}, 2018.

\bibitem{rnn}
D.~E. Rumelhart, G.~E. Hinton, and R.~J. Williams, ``Learning internal representations by error propagation, parallel distributed processing, explorations in the microstructure of cognition, ed. de rumelhart and j. mcclelland. vol. 1. 1986,'' \emph{Biometrika}, vol.~71, no. 599-607, p.~6, 1986.

\bibitem{gru}
D.~Bahdanau, K.~Cho, and Y.~Bengio, ``Neural machine translation by jointly learning to align and translate,'' \emph{arXiv preprint arXiv:1409.0473}, 2014.

\bibitem{lstm}
S.~Hochreiter and J.~Schmidhuber, ``Long short-term memory,'' \emph{Neural computation}, vol.~9, no.~8, pp. 1735--1780, 1997.

\bibitem{logtrans}
S.~Li, X.~Jin, Y.~Xuan, X.~Zhou, W.~Chen, Y.-X. Wang, and X.~Yan, ``Enhancing the locality and breaking the memory bottleneck of transformer on time series forecasting,'' \emph{Advances in neural information processing systems}, vol.~32, 2019.

\bibitem{reformer}
\BIBentryALTinterwordspacing
N.~Kitaev, L.~Kaiser, and A.~Levskaya, ``Reformer: The efficient transformer,'' in \emph{Intl. Conf. on Learning Representations}, 2020. [Online]. Available: \url{https://openreview.net/forum?id=rkgNKkHtvB}
\BIBentrySTDinterwordspacing

\bibitem{informer}
H.~Zhou, S.~Zhang, J.~Peng, S.~Zhang, J.~Li, H.~Xiong, and W.~Zhang, ``Informer: Beyond efficient transformer for long sequence time-series forecasting,'' in \emph{Proc. of the AAAI Conf. on artificial intelligence}, vol.~35, no.~12, 2021, pp. 11\,106--11\,115.

\bibitem{autoformer}
H.~Wu, J.~Xu, J.~Wang, and M.~Long, ``Autoformer: Decomposition transformers with auto-correlation for long-term series forecasting,'' \emph{Advances in Neural Information Processing Systems}, vol.~34, pp. 22\,419--22\,430, 2021.

\bibitem{fedformer}
T.~Zhou, Z.~Ma, Q.~Wen, X.~Wang, L.~Sun, and R.~Jin, ``{FEDformer}: Frequency enhanced decomposed transformer for long-term series forecasting,'' in \emph{Proc. 39th Intl. Conf. on Machine Learning (ICML 2022)}, 2022.

\bibitem{pyraformer}
S.~Liu, H.~Yu, C.~Liao, J.~Li, W.~Lin, A.~X. Liu, and S.~Dustdar, ``Pyraformer: Low-complexity pyramidal attention for long-range time series modeling and forecasting,'' in \emph{Intl. Conf. on Learning Representations}, 2022.

\bibitem{triformer}
R.-G. Cirstea, C.~Guo, B.~Yang, T.~Kieu, X.~Dong, and S.~Pan, ``Triformer: Triangular, variable-specific attentions for long sequence multivariate time series forecasting,'' in \emph{IJCAI}, 2022.

\bibitem{revin}
T.~Kim, J.~Kim, Y.~Tae, C.~Park, J.-H. Choi, and J.~Choo, ``Reversible instance normalization for accurate time-series forecasting against distribution shift,'' in \emph{Intl. Conf. on Learning Representations}, 2021.

\bibitem{vit}
\BIBentryALTinterwordspacing
A.~Dosovitskiy, L.~Beyer, A.~Kolesnikov, D.~Weissenborn, X.~Zhai, T.~Unterthiner, M.~Dehghani, M.~Minderer, G.~Heigold, S.~Gelly, J.~Uszkoreit, and N.~Houlsby, ``An image is worth 16x16 words: Transformers for image recognition at scale,'' in \emph{Intl. Conf. on Learning Representations}, 2021. [Online]. Available: \url{https://openreview.net/forum?id=YicbFdNTTy}
\BIBentrySTDinterwordspacing

\bibitem{pits}
S.~Lee, T.~Park, and K.~Lee, ``Learning to embed time series patches independently,'' \emph{arXiv preprint arXiv:2312.16427}, 2023.

\bibitem{mlpmixer}
I.~O. Tolstikhin, N.~Houlsby, A.~Kolesnikov, L.~Beyer, X.~Zhai, T.~Unterthiner, J.~Yung, A.~Steiner, D.~Keysers, J.~Uszkoreit \emph{et~al.}, ``{MLP-mixer: An all-MLP architecture for vision},'' \emph{Advances in neural information processing systems}, vol.~34, pp. 24\,261--24\,272, 2021.

\bibitem{resnet}
K.~He, X.~Zhang, S.~Ren, and J.~Sun, ``Deep residual learning for image recognition,'' in \emph{Proc. of the IEEE Conf. on computer vision and pattern recognition}, 2016, pp. 770--778.

\bibitem{layernorm}
J.~Lei~Ba, J.~R. Kiros, and G.~E. Hinton, ``Layer normalization,'' \emph{ArXiv e-prints}, pp. arXiv--1607, 2016.

\bibitem{cp}
A.~N. Angelopoulos and S.~Bates, ``A gentle introduction to conformal prediction and distribution-free uncertainty quantification,'' \emph{arXiv preprint arXiv:2107.07511}, 2021.

\bibitem{angelopoulos2020uncertainty}
A.~Angelopoulos, S.~Bates, J.~Malik, and M.~I. Jordan, ``Uncertainty sets for image classifiers using conformal prediction,'' \emph{arXiv preprint arXiv:2009.14193}, 2020.

\bibitem{papadopoulos2008inductive}
H.~Papadopoulos, ``Inductive conformal prediction: Theory and application to neural networks,'' in \emph{Tools in artificial intelligence}.\hskip 1em plus 0.5em minus 0.4em\relax Citeseer, 2008.

\bibitem{vovk1999machine}
V.~Vovk, A.~Gammerman, and C.~Saunders, ``Machine-learning applications of algorithmic randomness,'' \emph{PMLR}, 1999.

\bibitem{adda2023highnoon}
S.~Adda, L.~Chiaraviglio, D.~Franci, C.~Lodovisi, N.~Pasquino, S.~Pavoncello, C.~Pedroli, and R.~Pelosini, ``High noon for mobile networks: Short-time {EMF} measurements to capture daily exposure,'' \emph{IEEE Trans. on Instrumentation and Measurement}, vol.~72, pp. 1--10, 2023.

\bibitem{adam}
D.~P. Kingma and J.~Ba, ``Adam: A method for stochastic optimization,'' \emph{arXiv preprint arXiv:1412.6980}, 2014.

\bibitem{tsmixer}
S.-A. Chen, C.-L. Li, N.~Yoder, S.~O. Arik, and T.~Pfister, ``{TSmixer: An all-MLP} architecture for time series forecasting,'' \emph{arXiv preprint arXiv:2303.06053}, 2023.

\end{thebibliography}
\bibliographystyle{IEEEtran}

\end{document}